\documentclass{article}

\makeatletter
\def\input@path{%
 {./}%
 {docs/iclr2027/}%
 {sda-xarch-transfer/docs/iclr2027/}%
 {../iclr2027/}%
}
\makeatother

\usepackage{iclr2027_conference,times}
\iclrfinalcopy 

\usepackage{amsmath,amsfonts,amssymb,bm}

\def\eqref#1{equation~\ref{#1}}

\def\1{\bm{1}}

\DeclareMathAlphabet{\mathsfit}{\encodingdefault}{\sfdefault}{m}{sl}
\SetMathAlphabet{\mathsfit}{bold}{\encodingdefault}{\sfdefault}{bx}{n}

\usepackage[T1]{fontenc}
\usepackage{hyperref}
\usepackage{url}
\usepackage{booktabs}
\usepackage{tabularx}
\usepackage{flafter} 
\usepackage{array}
\usepackage{ragged2e}
\usepackage{makecell}
\usepackage{xcolor}
\usepackage{colortbl}
\usepackage{graphicx}  
\usepackage{tikz}
\usetikzlibrary{arrows.meta,positioning,calc,fit,backgrounds,matrix,decorations.pathreplacing}
\usepackage{microtype}
\usepackage{placeins}
\usepackage{float} 
\usepackage{capt-of} 
\usepackage{enumitem}
\setlist[itemize]{leftmargin=*,itemsep=0pt,parsep=0pt,topsep=2pt,partopsep=0pt}
\setlist[enumerate]{leftmargin=*,itemsep=0pt,parsep=0pt,topsep=2pt,partopsep=0pt}

\definecolor{evalcls}{RGB}{180,30,30}
\definecolor{outcls}{RGB}{0,90,140}
\definecolor{cellok}{RGB}{220,245,220}
\definecolor{cellbad}{RGB}{255,230,230}
\newcommand{\disc}[1]{#1}
\newcommand{\dischl}[1]{#1}

\newcommand{\steerNA}{\textcolor{outcls}{\textbf{N/A}}}
\newcommand{\statusEff}{effective}
\newcommand{\statusIneff}{ineffective}

\newcommand{\statusInfeas}{infeasible}
\newcommand{\ool}{\ensuremath{^{\mathrm{ol}}}}
\setcellgapes{0pt}
\newcolumntype{Z}{@{}%
 >{\RaggedRight\arraybackslash}p{0.18\linewidth}%
 >{\RaggedRight\arraybackslash}p{0.20\linewidth}%
 >{\RaggedRight\sloppy\arraybackslash}X@{}}
\newcommand{\siteclassmaprestablesetup}{%
 \siteclassmaptablesetup
}
\newcommand{\siteclassmapresheader}{%
  Class & Contrast / site & Status (why / lim) \\
}
\newcommand{\scwl}[3]{%
  \makecell[tl]{#1~\textbf{why:}~#2\\[-0.08em]\textbf{lim:}~#3}%
}
\newcommand{\siteclassmapfootnote}{%
 {\scriptsize\RaggedRight
 $^*$Guard: label miss. $^\ddagger$MTP / LM$+$DiT: \steerNA{}.
  \ool~out-of-lineage. s$+$c$=$self$+$cross.
  Steer-N/A$=$no residual hook (vs.\ \statusIneff{}/\statusInfeas{}).
  $\mathrm{R}_{\mathrm{cond}}$ / ABL\_POS$=$defence role (not attack).}%
}
\newcommand{\siteclassmaptablesetup}{%
 \scriptsize
 \setlength{\tabcolsep}{1.2pt}%
 \renewcommand{\arraystretch}{0.80}%
 \setlength{\extrarowheight}{0pt}%
}

\title{Where Is Rank-1 DiM Refusal Intervenable in Multimodal LM Derivatives?}

\author{
Chansu Han \\
NICT \\
\texttt{han@nict.go.jp}
\And
Seok-Hwan Choi \\
Yonsei University
\And
Takeshi Takahashi \\
NICT
}

\begin{document}
\maketitle

\begin{abstract}
Open-weight model hubs now ship not only chat checkpoints but a long tail of
derivatives---fine-tunes, safety guards, vision--language models (VLMs), audio
wrappers, and generative heads---yet residual-stream safety audits still often
intervene at a single site, usually the text decoder~$T$.
We ask a localization question under one shared mean / rank-1
difference-in-means (DiM) extract$=$intervene protocol, with matched-norm random
and empty-rate controls: does causal chat attack lift concentrate at~$T$, or do
perceptual encoders, projectors, and fusion paths ($E$/$P$/$F$) supply a
comparable bypass?
On the mixed-input settings we stress-test---HADES adversarial images paired
with harmful text prompts, and spoken AdvBench via text-to-speech (TTS) or
silence---text DiM at~$T$ raises attack success rate (ASR) while the same mean
DiM at $E$/$P$/$F$ does not; attacks track harmful text rather than image or
audio content.
Margin-rank and leverage probes on a Vision leaf show that this recipe is
extremal at~$T$ and near chance at $P$/$E$, which pressures a weak-extractor
reading of those flat cells for the DiM family we study.
Catalog breadth and mapper-free base$\to$derivative transfer when residual widths
match provide supporting coverage under the same gates; generative U-Nets are a
separate band under generation judges.
\disc{The primary claim is this recipe-scoped $T$ vs.\ non-text intervenability
asymmetry---intervention accessibility under mean / rank-1 DiM, not a claim that
refusal is absent or fully represented outside~$T$ (text vs.\ perceptual
contrasts and pooling differ by site).
On VLGuard Unsafe prompts ($n{=}200$; Llama-Guard), the same $T$ vs.\ $P$/$E$
directionality holds at modest absolute ASR
(\dischl{{$0.030{\to}0.150$}} at~$T$).
An intervention-oriented Class$\times$site ledger records where this protocol still
hooks as supporting coverage, not as a universal multimodal bypass.}
\end{abstract}

\section{Introduction}
\label{sec:intro}
Open-weight hubs ship not only official instruct checkpoints~$B$, but a long
tail of derivatives~$M$: continued fine-tunes, safety guards, thinking models,
vision--language models (VLMs), audio wrappers, mixer hybrids, and speculative
or draft heads.
Hub cards rarely document which residual stream an auditor can still hook, or
whether a difference-in-means (DiM) refusal direction
\citep{arditi2024refusal,marshall2024affine} remains causally effective after
that change.
Product-level ontologies and fine-tuning risk surveys
\citep{qi2024finetuning,huang2026ftsurvey} describe \emph{what} was trained;
they do not tell an intervention auditor \emph{where} to extract and ablate.
Single-model refusal DiM and same-architecture vector reuse
\citep{arditi2024refusal,zhu2024safetylock,wu2026enchtable} stress one
checkpoint or transfer a defence, while learned cross-model maps
\citep{oozeer2025activation} ask whether an intervention can be \emph{moved}.
To our knowledge, prior work has not jointly (i)~asked, under one shared mean /
rank-1 DiM extract$=$intervene protocol with matched-norm random controls,
whether causal chat attack success rate (ASR) lift concentrates at~$T$ or also
arises at perceptual encoders, projectors, and fusion ($E$/$P$/$F$),
(ii)~catalogued open-weight derivative Classes that span continued fine-tunes,
guards, thinking and speculative heads, VLMs, audio wrappers, and mixer hybrids
and audited each under that same protocol, and (iii)~reported mapper-free
base$\to$derivative cross for base / self / cross donors on the same gates when
residual widths match---with (ii)--(iii) as supporting coverage of where the
protocol still hooks, not as the primary localization claim.

We therefore adopt a compact, intervention-oriented taxonomy
(Root / Inheritance / Axis / Class; Figure~\ref{fig:class-site-grid}a)
and map each Class to residual hook sites
(Table~\ref{apptab:site-class-map}; extraction IDs in
Table~\ref{tab:dim-recipe}).
The operative question is not a full language-model ontology, but which
architectural change breaks the \emph{same} rank-1 residual protocol
\citep{oozeer2025activation}.

\textbf{Main question.}
Under one shared mean / rank-1 DiM protocol, does causal chat attack lift
concentrate at the text decoder~$T$, or do perceptual encoders ($E$), projectors
($P$), and fusion paths ($F$) supply comparable intervenability?
\disc{We answer this as a question about \emph{intervention accessibility under this
recipe family}: text DiM and non-text DiM change the contrast variable and
pooling with the site, so the comparison is not a fully isolated location effect.}

\textbf{Findings.}
(i)~\emph{Chat attack lift is $T$-local on the cells we test.}
Text DiM at~$T$ raises chat ASR on plain text and on tested mixed-input wrappers
(HADES adversarial image$+$harmful text; Qwen2-Audio TTS or silence), while mean
non-text DiM $\mathrm{R}_{\mathrm{u}}$ at $E$/$P$/$F$ yields no attack lift under
matched-norm random.
A cross-condition with a benign image and harmful text shows that attacks track
harmful text, not image content (typography-style FigStep remains a ceiling;
Sec.~\ref{sec:emp}).
\disc{HADES-style cells are therefore stress tests of $T$-side intervenability under
multimodal \emph{inputs}, not maximally diagnostic modality-native refusal;
VLGuard Unsafe ($n{=}200$; Llama-Guard) preserves $T{>}P$/$E$ directionality at
modest absolute ASR (Sec.~\ref{sec:emp}).}
On the Vision-Instruct leaf that carries the HADES win, the extracted direction
at~$T$ is worst-case among 50 matched-norm margin probes, whereas at $P$/$E$ it
is near chance (Figure~\ref{fig:mechanism})---evidence that the tested mean /
rank-1 recipe is strong at~$T$ and weak at perception sites, without claiming
that no other feature family could succeed there.
(ii)~\emph{Supporting coverage under the same gates.}
Llama-lineage~$T$ is broadly effective, with mapper-free $B\!\to\!M$ cross when
widths match; Thinking models are Class-effective only with suite qualification.
Hard-Class probes are deferred to Appendix~\ref{app:fair-tables}.
(iii)~\emph{Generative U-Nets are a separate band.}
Concept DiM erases harm at down-blocks and fails at mid-blocks under generation
judges we never pool with chat ASR.

\textbf{Contributions.}
(1)~A shared extract$=$intervene audit that applies Arditi-style mean DiM at both
$T$ and non-text sites under matched-norm random, and reports the resulting
$T$ vs.\ $E$/$P$/$F$ asymmetry---including the negative non-text result for this
recipe family---with margin-rank and leverage instruments on a Vision leaf
(\S\ref{sec:mech}).
(2)~An intervention-oriented Class taxonomy of open-weight derivatives with
audited per-Class coverage under one protocol.
(3)~Mapper-free supply-chain cross when residual widths match, together with
within-protocol recipe comparisons (Fig.~\ref{fig:method-compare}), as supporting
coordinates rather than as the primary localization claim.

\section{Related work}
\label{sec:rw}
\textbf{Refusal geometry.}
Arditi et~al.\ extract a single DiM direction whose ablation or
orthogonalization suppresses refusal in-model \citep{arditi2024refusal}.
Yu et~al.\ keep the same DiM axis but replace zeroing with refusal-feature ablation
(RFA)---patching the
refusal (or negative-side) coordinate to $(\mu^{(-)\top}\hat{v})\,\hat{v}$---as a
proxy worst-case safety offset used inside adversarial training \citep{yu2025refat};
we record that intervention (not ReFAT training) as an optional recipe on any
$(\mathrm{R}_{\ast},\mathrm{site})$ pair in Sec.~\ref{sec:nt-dim}.
Follow-ups refine the extractor \citep{siu2025cosmic}, argue for multi-direction
or affine structure \citep{joad2026multidir,marshall2024affine}, and separate
harmfulness from refusal features \citep{zhao2025harmrefuse}.
We keep Arditi-style rank-1 DiM as the shared substrate so taxonomy Class is about
\emph{whether the same protocol still hooks}, not about a new extractor;
RFA is tested as a within-Class recipe, not a taxonomy change.

\textbf{Transfer of interventions.}
SafetyLock and EnchTable reuse safety vectors across same-architecture
fine-tunes \citep{zhu2024safetylock,wu2026enchtable};
Oozeer et~al.\ transfer activation interventions via \emph{learned} maps
\citep{oozeer2025activation};
Agarwal studies sparse autoencoder (SAE)$+$bridge cross-architecture steering
\citep{agarwal2026crosssteer}.
Those lines ask whether a defence or steer can be \emph{moved}.
We instead ask, for each derivative Class, whether self DiM at a documented site
is an attack \statusEff{}---and, when widths match, whether a family / instruct-base
donor $\hat{v}_B$ cross-loads without a mapper
($\mathrm{ASR}_{\mathrm{base}}$/$\mathrm{ASR}_{\mathrm{self}}$/$\mathrm{ASR}_{\mathrm{cross}}$).

\textbf{Multimodal residual audits.}
Yin et~al.\ audit image-conditioning / ABL\_POS directions at the VLM decoder as
inference-time defence \citep{yin2026crossarch}.
MARS harnesses textual refusal directions for multimodal safety inside one VLM
\citep{dinca2026mars}.
ASTRA attributes over visual tokens but still steers at~$T$ \citep{wang2025astra}.
CMRM / ShiftDC calibrate modality-induced decoder shifts as defence
\citep{liu2025cmrm,zou2025shiftdc}.
HADES supplies adversarial images as generation inputs \citep{li2024hades}.
Across these lines, extract or intervene typically remains at the text decoder;
non-text activations are used mainly for attribution or conditioning, not as
same-site attack DiM sites.
We use HADES and spoken AdvBench as mixed-input \emph{eval} cells with intervene@$T$,
and additionally audit same-site mean $\mathrm{R}_{\mathrm{u}}$@$E$/$P$/$F$ under
matched-norm random (Sec.~\ref{sec:emp}), treating a flat non-text outcome as
evidence about this recipe family rather than as a failed experiment.
Beyond protocol axes (App.~\ref{app:rw-map}; Table~\ref{tab:related-compare}),
the Vision same-pack residual bake-off is Fig.~\ref{fig:method-compare}a
(Sec.~\ref{sec:setup}); LLaVA footnotes are Table~\ref{apptab:asr-bakeoff}.
We do not paste published unsafe rates next to attack ASR, and we do not
reimplement full ASTRA token-attribution as an alternate attack recipe.

\textbf{Benchmarks and supply-chain risk.}
AdvBench \citep{zou2023advbench}, HarmBench \citep{mazeika2024harmbench}, and
StrongREJECT \citep{souly2024strongreject} standardize harmful requests;
Llama Guard scores labels rather than chat completions \citep{inan2023llamaguard}.
Fine-tuning can erase alignment \citep{qi2024finetuning,huang2026ftsurvey}.
Dedicated taxonomies of architecture-variant open-weight derivatives, as an
\emph{intervention} axis under a shared extract$=$intervene protocol with
mapper-free $B\!\to\!M$ cross, remain thin relative to FT-risk surveys and to
single-model DiM / mapper-transfer studies (above)---the coverage gap our
supporting Class ledger fills around the $T$ vs.\ non-text localization claim.

\section{Derivative taxonomy}
\label{sec:taxonomy}

\textbf{Role.}
Intervention-oriented labeling for residual-stream audits (not a complete LM ontology).
Quantization, context length, and SFT/DPO/abliteration are model attributes, not new Axes.
In figures, \textcolor{evalcls}{red} marks Classes that are \statusEff{} under our
gates (including Guard label miss$\uparrow$; suite- or $\lambda$-limited lifts that
still clear matched-norm random count as red); \textcolor{outcls}{blue} marks Classes
that remain \statusIneff{}, \statusInfeas{}, or \steerNA{}.

\paragraph{Outcome labels.}
\begin{itemize}
\item \statusEff{}---steered ASR (or task-native erase/miss) above base with
matched-norm random near floor and non-collapsed generations; suite- or
narrow-$\lambda$ lifts that clear those gates stay Class-\statusEff{} (red).
\item \statusIneff{}---documented site runs, but no gated attack lift
(ASR flat, anti-lift, or random$\approx$self).
\item \statusInfeas{}---same-band checkpoint exists, yet loader, hook, tokenizer,
or eval saturation blocks a fair run.
\item \steerNA{}---no causal residual hook on the scored path (non-AR forward,
action/VLA readout, MTP/draft-only, bit kernels, pure SSM without exposed
$h^{(\ell)}$)---applicability limit, not a failed search.
\end{itemize}
Where no Llama-lineage checkpoint covers an Axis, we evaluate an out-of-lineage
Hub model (marked~\ool; self DiM; within-family cross when widths match).
Chat \statusEff{}/\statusIneff{} follow Sec.~\ref{sec:setup} gates
(Table~\ref{apptab:site-class-map}); Fig.~\ref{fig:mechanism} shows the tested
mean / rank-1 recipe strong at Vision~$T$ and near chance at $P$/$E$.
Hook layout: App.~\ref{app:hook-diagram}.

\textbf{Levels \& band.}
Root $\to$ Inheritance $\to$ Axis (A1--A7) $\to$ Class.
Intervention band (Same-shape / Resized / Multimodal / Mixer) marks how easily the
same rank-1 residual protocol runs as on~$B$; A3/A7 may be Band~---.
Eval Root is Llama-3.1-8B-Instruct (primary lineage for Same-band and Llama-backed
wrappers).
Cross uses a within-family donor (Llama $B\!\to\!M$; width-matched Qwen
$B'\!\to\!M$).

\begin{figure}[!t]
\centering
\resizebox{\linewidth}{!}{%
\tikzset{
  hdr/.style={font=\scriptsize\sffamily\bfseries, text=black!75, align=center},
  blk/.style={font=\scriptsize\sffamily\bfseries, text=black!80, anchor=west},
  axl/.style={font=\scriptsize\sffamily\bfseries, text=black!62},
  cls/.style={font=\scriptsize, anchor=west, align=left, inner sep=0pt},
  mdl/.style={font=\scriptsize\itshape, anchor=west, align=left, inner sep=0pt, text=black!52},
  bandt/.style={font=\tiny\sffamily, text=black!58, align=center},
  pnl/.style={font=\small\bfseries, anchor=west},
  gloss/.style={font=\tiny\itshape, text=black!50},
  cell/.style={minimum width=0.68cm, minimum height=0.245cm, inner sep=0.4pt,
               font=\tiny\bfseries, align=center, rounded corners=0.6pt},
  eff/.style ={cell, fill=evalcls!16, draw=evalcls!70, text=evalcls},
  inef/.style={cell, fill=outcls!12,  draw=outcls!65,  text=outcls},
  part/.style={cell, fill=outcls!6,   draw=outcls!55, densely dashed, text=outcls},
  infe/.style={cell, fill=black!7,    draw=black!35,   text=black!55},
  nav/.style ={cell, fill=black!4,    draw=black!25,   text=black!45},
  ntlab/.style={font=\tiny, anchor=west, text=black!58, inner sep=0pt},
  scell/.style={cell, minimum width=0.90cm, minimum height=0.38cm},
  bandhdr/.style={fill=black!5, rounded corners=1.5pt, inner xsep=5pt, inner ysep=2.5pt,
                  font=\tiny\sffamily, text=black!70},
}
\begin{tikzpicture}[x=1cm,y=1cm]
\node[pnl] at (-0.15,1.05) {(a) Which residual site is intervenable, per Class};
\node[blk] at (-0.15,0.68) {Axes whose only causal residual site is~$T$};
\node[hdr, anchor=west] at (0,0.32) {Class \textit{(model)}};
\node[hdr] at (4.08,0.32) {Band};
\node[hdr] at (5.04,0.32) {$T$};
\draw[black!45, line width=0.5pt] (-0.15,0.12) -- (5.54,0.12);
\node[font=\tiny\itshape, text=black!48] at (4.08,0.02) {vs.\ instruct root~$B$};
\node[axl, anchor=west] at (-0.15,-0.385) {Axis A1~~tuning / compression};
\node[cls, text=evalcls] at (0,-0.720) {Instruct root};
\node[mdl] at (1.8,-0.720) {Llama-8B};
\node[bandt] at (4.08,-0.720) {Same};
\node[eff] at (5.04,-0.720) {E};
\node[cls, text=evalcls] at (0,-1.055) {Wild FT};
\node[mdl] at (1.8,-1.055) {Hermes, Tulu};
\node[bandt] at (4.08,-1.055) {Same};
\node[eff] at (5.04,-1.055) {E};
\node[cls, text=evalcls] at (0,-1.390) {Domain FT};
\node[mdl] at (1.8,-1.390) {UltraMedical};
\node[bandt] at (4.08,-1.390) {Same};
\node[eff] at (5.04,-1.390) {E};
\node[cls, text=evalcls] at (0,-1.725) {Size variant};
\node[mdl] at (1.8,-1.725) {Llama-3.2-3B};
\node[bandt] at (4.08,-1.725) {Resized};
\node[eff] at (5.04,-1.725) {E};
\node[cls, text=evalcls] at (0,-2.060) {Distill};
\node[mdl] at (1.8,-2.060) {R1, SN};
\node[bandt] at (4.08,-2.060) {Same};
\node[eff] at (5.04,-2.060) {E};
\node[cls, text=evalcls] at (0,-2.395) {PEFT / LoRA};
\node[mdl] at (1.8,-2.395) {S5};
\node[bandt] at (4.08,-2.395) {Same};
\node[eff] at (5.04,-2.395) {E};
\node[cls, text=evalcls] at (0,-2.730) {Official MoE};
\node[mdl] at (1.8,-2.730) {Mixtral\ool};
\node[bandt] at (4.08,-2.730) {Resized};
\node[eff] at (5.04,-2.730) {E};
\node[cls, text=outcls] at (0,-3.065) {Compress};
\node[mdl] at (1.8,-3.065) {Sheared};
\node[bandt] at (4.08,-3.065) {Same/Rsz};
\node[inef] at (5.04,-3.065) {I};
\node[cls, text=outcls] at (0,-3.400) {Compress};
\node[mdl] at (1.8,-3.400) {Qwen-prune\ool};
\node[bandt] at (4.08,-3.400) {Same};
\node[inef] at (5.04,-3.400) {I};
\node[axl, anchor=west] at (-0.15,-3.995) {Axis A2~~decoding \& safety heads};
\node[cls, text=evalcls] at (0,-4.330) {Guard};
\node[mdl] at (1.8,-4.330) {Guard-3};
\node[bandt] at (4.08,-4.330) {Same};
\node[eff] at (5.04,-4.330) {E};
\node[cls, text=evalcls] at (0,-4.665) {Thinking};
\node[mdl] at (1.8,-4.665) {R1-Llama};
\node[bandt] at (4.08,-4.665) {Same};
\node[eff] at (5.04,-4.665) {E};
\node[cls, text=outcls] at (0,-5.000) {Thinking};
\node[mdl] at (1.8,-5.000) {R1-Qwen\ool};
\node[bandt] at (4.08,-5.000) {Same};
\node[inef] at (5.04,-5.000) {I};
\node[cls, text=outcls] at (0,-5.335) {Speculative};
\node[mdl] at (1.8,-5.335) {EAGLE3};
\node[bandt] at (4.08,-5.335) {---};
\node[infe] at (5.04,-5.335) {X};
\node[cls, text=outcls] at (0,-5.670) {MTP head};
\node[mdl] at (1.8,-5.670) {V3\ool};
\node[bandt] at (4.08,-5.670) {---};
\node[nav] at (5.04,-5.670) {N/A};
\node[axl, anchor=west] at (-0.15,-6.265) {Axis A3~~training objective};
\node[cls, text=outcls] at (0,-6.600) {AR$\to$diffusion};
\node[mdl] at (1.8,-6.600) {Dream\ool, LLaDA};
\node[bandt] at (4.08,-6.600) {---};
\node[inef] at (5.04,-6.600) {I};
\node[blk] at (6.3999999999999995,0.68) {Axes that expose a non-$T$ site};
\node[hdr, anchor=west] at (6.55,0.32) {Class \textit{(model)}};
\node[hdr] at (10.629999999999999,0.32) {Band};
\node[hdr] at (11.59,0.32) {$T$};
\node[hdr, anchor=west] at (12.09,0.32) {non-$T$ site};
\draw[black!45, line width=0.5pt] (6.3999999999999995,0.12) -- (14.78,0.12);
\node[axl, anchor=west] at (6.3999999999999995,-0.385) {Axis A4~~multimodal wrappers \& heads};
\node[cls, text=evalcls] at (6.55,-0.720) {Official VLM};
\node[mdl] at (8.35,-0.720) {Vision-Instr.};
\node[bandt] at (10.629999999999999,-0.720) {MM};
\node[eff] at (11.59,-0.720) {E};
\node[inef] at (12.43,-0.720) {I};
\node[ntlab] at (12.83,-0.720) {$E$/$P$/$F$/$C$};
\node[cls, text=evalcls] at (6.55,-1.055) {Comm.\ VLM};
\node[mdl] at (8.35,-1.055) {LLaVA-Mini};
\node[bandt] at (10.629999999999999,-1.055) {MM};
\node[eff] at (11.59,-1.055) {E};
\node[inef] at (12.43,-1.055) {I};
\node[ntlab] at (12.83,-1.055) {$\mathrm{R}_\mathrm{u}$ sites};
\node[cls, text=evalcls] at (6.55,-1.390) {Audio wrapper};
\node[mdl] at (8.35,-1.390) {Qwen2-Audio\ool};
\node[bandt] at (10.629999999999999,-1.390) {MM};
\node[eff] at (11.59,-1.390) {E};
\node[inef] at (12.43,-1.390) {I};
\node[ntlab] at (12.83,-1.390) {projector};
\node[cls, text=evalcls] at (6.55,-1.725) {Audio wrapper};
\node[mdl] at (8.35,-1.725) {DeSTA2.5 (RFA)};
\node[bandt] at (10.629999999999999,-1.725) {MM};
\node[eff] at (11.59,-1.725) {E};
\node[cls, text=evalcls] at (6.55,-2.060) {Native MM};
\node[mdl] at (8.35,-2.060) {Omni\ool};
\node[bandt] at (10.629999999999999,-2.060) {MM};
\node[eff] at (11.59,-2.060) {E};
\node[cls, text=evalcls] at (6.55,-2.395) {Gen.\ MM head};
\node[mdl] at (8.35,-2.395) {SD-T\ool, ALD\ool};
\node[bandt] at (10.629999999999999,-2.395) {---};
\node[eff] at (12.43,-2.395) {E};
\node[ntlab] at (12.83,-2.395) {UNet dn\,E / mid\,I};
\node[axl, anchor=west] at (6.3999999999999995,-2.990) {Axis A5~~sequence mixer};
\node[cls, text=evalcls] at (6.55,-3.325) {Hybrid mixer};
\node[mdl] at (8.35,-3.325) {Mamba2};
\node[bandt] at (10.629999999999999,-3.325) {Mixer};
\node[inef] at (11.59,-3.325) {I};
\node[eff] at (12.43,-3.325) {E};
\node[ntlab] at (12.83,-3.325) {Xm/Xa};
\node[cls, text=evalcls] at (6.55,-3.660) {Native mixer};
\node[mdl] at (8.35,-3.660) {RecGemma\ool};
\node[bandt] at (10.629999999999999,-3.660) {Mixer};
\node[eff] at (11.59,-3.660) {E};
\node[cls, text=evalcls] at (6.55,-3.995) {Dense$\to$MoE};
\node[mdl] at (8.35,-3.995) {Qwen1.5\ool};
\node[bandt] at (10.629999999999999,-3.995) {Resized};
\node[eff] at (11.59,-3.995) {E};
\node[cls, text=outcls] at (6.55,-4.330) {Attn replace};
\node[mdl] at (8.35,-4.330) {DS-V2\ool};
\node[bandt] at (10.629999999999999,-4.330) {Mixer};
\node[inef] at (11.59,-4.330) {I};
\node[cls, text=outcls] at (6.55,-4.665) {Pure SSM};
\node[mdl] at (8.35,-4.665) {Falcon-M\ool};
\node[bandt] at (10.629999999999999,-4.665) {Mixer};
\node[nav] at (12.43,-4.665) {N/A};
\node[ntlab] at (12.83,-4.665) {mixer};
\node[axl, anchor=west] at (6.3999999999999995,-5.260) {Axis A6/A7~~agents \& numeric format};
\node[cls, text=evalcls] at (6.55,-5.595) {Thin agent};
\node[mdl] at (8.35,-5.595) {ToolACE, xLAM};
\node[bandt] at (10.629999999999999,-5.595) {---};
\node[eff] at (11.59,-5.595) {E};
\node[cls, text=outcls] at (6.55,-5.930) {VLA};
\node[mdl] at (8.35,-5.930) {MolmoAct\ool};
\node[bandt] at (10.629999999999999,-5.930) {MM};
\node[nav] at (11.59,-5.930) {N/A};
\node[cls, text=outcls] at (6.55,-6.265) {LM$+$DiT};
\node[mdl] at (8.35,-6.265) {$\pi$0, CogACT\ool};
\node[bandt] at (10.629999999999999,-6.265) {MM};
\node[nav] at (11.59,-6.265) {N/A};
\node[cls, text=outcls] at (6.55,-6.600) {BitNet};
\node[mdl] at (8.35,-6.600) {b1.58\ool};
\node[bandt] at (10.629999999999999,-6.600) {---};
\node[nav] at (11.59,-6.600) {N/A};
\draw[evalcls!55, line width=0.7pt, rounded corners=1.5pt] (4.650,-0.560) rectangle (5.430,-6.760);
\draw[evalcls!55, line width=0.7pt, rounded corners=1.5pt] (11.200,-0.560) rectangle (11.980,-6.760);
\node[pnl] at (-0.15,-7.020) {(b) Which input condition, per site};
\node[bandhdr] at (4.355,-7.580) {scored output $=$ \textbf{text} (chat judges)};
\node[align=center, font=\tiny] at (2.780,-8.080) {$T_H$\\[-0.15em]{\tiny\itshape harmful text}};
\node[align=center, font=\tiny] at (3.830,-8.080) {$I{+}T_H$\\[-0.15em]{\tiny\itshape $+$image}};
\node[align=center, font=\tiny] at (4.880,-8.080) {$I_B{+}T_H$\\[-0.15em]{\tiny\itshape $+$benign img}};
\node[align=center, font=\tiny] at (5.930,-8.080) {$A{+}T_H$\\[-0.15em]{\tiny\itshape $+$audio}};
\node[anchor=east, font=\scriptsize] at (2.260,-8.600) {$E$~{\tiny\itshape encoder}};
\node[scell,inef] at (3.830,-8.600) {I};
\node[anchor=east, font=\scriptsize] at (2.260,-9.020) {$P$~{\tiny\itshape projector}};
\node[scell,inef] at (3.830,-9.020) {I};
\node[scell,inef] at (5.930,-9.020) {I};
\node[anchor=east, font=\scriptsize] at (2.260,-9.440) {$F$~{\tiny\itshape fusion}};
\node[scell,inef] at (3.830,-9.440) {I};
\node[anchor=east, font=\scriptsize] at (2.260,-9.860) {$T$~{\tiny\itshape text decoder}};
\node[scell,eff] at (2.780,-9.860) {E};
\node[scell,eff] at (3.830,-9.860) {E};
\node[scell,eff] at (4.880,-9.860) {E};
\node[scell,eff] at (5.930,-9.860) {E};
\node[bandhdr] at (9.500,-7.580) {scored output $=$ \textbf{image / audio} (erase judges)};
\node[align=center, font=\tiny] at (8.450,-8.080) {$T{\to}I$\\[-0.15em]{\tiny\itshape text$\to$img}};
\node[align=center, font=\tiny] at (9.500,-8.080) {$I{\to}I$\\[-0.15em]{\tiny\itshape img$\to$img}};
\node[align=center, font=\tiny] at (10.550,-8.080) {$T{\to}A$\\[-0.15em]{\tiny\itshape text$\to$audio}};
\node[anchor=east, font=\scriptsize] at (7.930,-8.600) {UNet-dn};
\node[scell,eff] at (8.450,-8.600) {E};
\node[scell,eff] at (9.500,-8.600) {E};
\node[scell,eff] at (10.550,-8.600) {E};
\node[anchor=east, font=\scriptsize] at (7.930,-9.020) {UNet-mid};
\node[scell,inef] at (8.450,-9.020) {I};
\node[scell,inef] at (10.550,-9.020) {I};
\node[anchor=south, rotate=90, font=\tiny\sffamily, text=black!45] at (0.60,-9.250) {intervention site};
\draw[evalcls!70, line width=0.7pt, rounded corners=1.5pt] (2.260,-10.090) rectangle (6.450,-9.630);
\node[anchor=north west, font=\tiny, text=evalcls!80!black] at (2.260,-10.160) {$T$ holds under every input condition};
\draw[evalcls!70, line width=0.7pt, rounded corners=1.5pt] (7.930,-8.830) rectangle (11.070,-8.370);
\node[anchor=north west, font=\tiny, text=evalcls!80!black] at (7.430,-9.300) {UNet-dn holds under every generative condition};
\node[anchor=north west, font=\tiny\itshape, text=black!50] at (7.430,-9.540) {dn / mid $=$ U-Net down / mid blocks};
\node[eff, minimum width=0.56cm] at (11.95,-8.020) {E};
\node[font=\tiny\bfseries, anchor=west, text=black!70] at (12.309999999999999,-8.020) {effective};
\node[inef, minimum width=0.56cm] at (11.95,-8.330) {I};
\node[font=\tiny\bfseries, anchor=west, text=black!70] at (12.309999999999999,-8.330) {ineffective};
\node[infe, minimum width=0.56cm] at (11.95,-8.640) {X};
\node[font=\tiny\bfseries, anchor=west, text=black!70] at (12.309999999999999,-8.640) {infeasible};
\node[font=\tiny, anchor=west, text=black!55] at (12.309999999999999,-8.900) {loader / tokenizer / eval blocked};
\node[nav, minimum width=0.56cm] at (11.95,-9.180) {N/A};
\node[font=\tiny\bfseries, anchor=west, text=black!70] at (12.309999999999999,-9.180) {Steer-N/A};
\node[font=\tiny, anchor=west, text=black!55] at (12.309999999999999,-9.440) {no residual hook on scored path};
\node[font=\tiny, anchor=north west, text=black!62, align=left] at (-0.15,-10.480) {Class name \textcolor{evalcls}{red} $=$ Class-effective, \textcolor{outcls}{blue} $=$ ineffective / infeasible / Steer-N/A; \ool~out-of-lineage (self DiM; family cross when width matches).\\
Empty cell $=$ not claimed. Band $=$ width/template relation to instruct root~$B$.};
\end{tikzpicture}

}
\caption{\textbf{Class$\times$site map} under one extract$=$intervene protocol
(taxonomy overview; not an ASR plot).
\textbf{(a)}~Per-Class intervenable residual site (non-$T$ drawn only when present).
Band $=$ width/template relation to the instruct root~$B$.
Llama-lineage~$T$ is broadly \statusEff{}; existing non-$T$ sites stay
\statusIneff{} except UNet-down and the Mamba2 mixer.
\textbf{(b)}~$T$ holds under all four chat input conditions; $E$/$P$/$F$ stay
\statusIneff{}; UNet-dn holds on every generative condition, mid does not
(judges never pooled across bands).
Numbers: Table~\ref{tab:localization}; per-Class why/lim:
Table~\ref{apptab:site-class-map}.}
\label{fig:class-site-grid}
\end{figure}
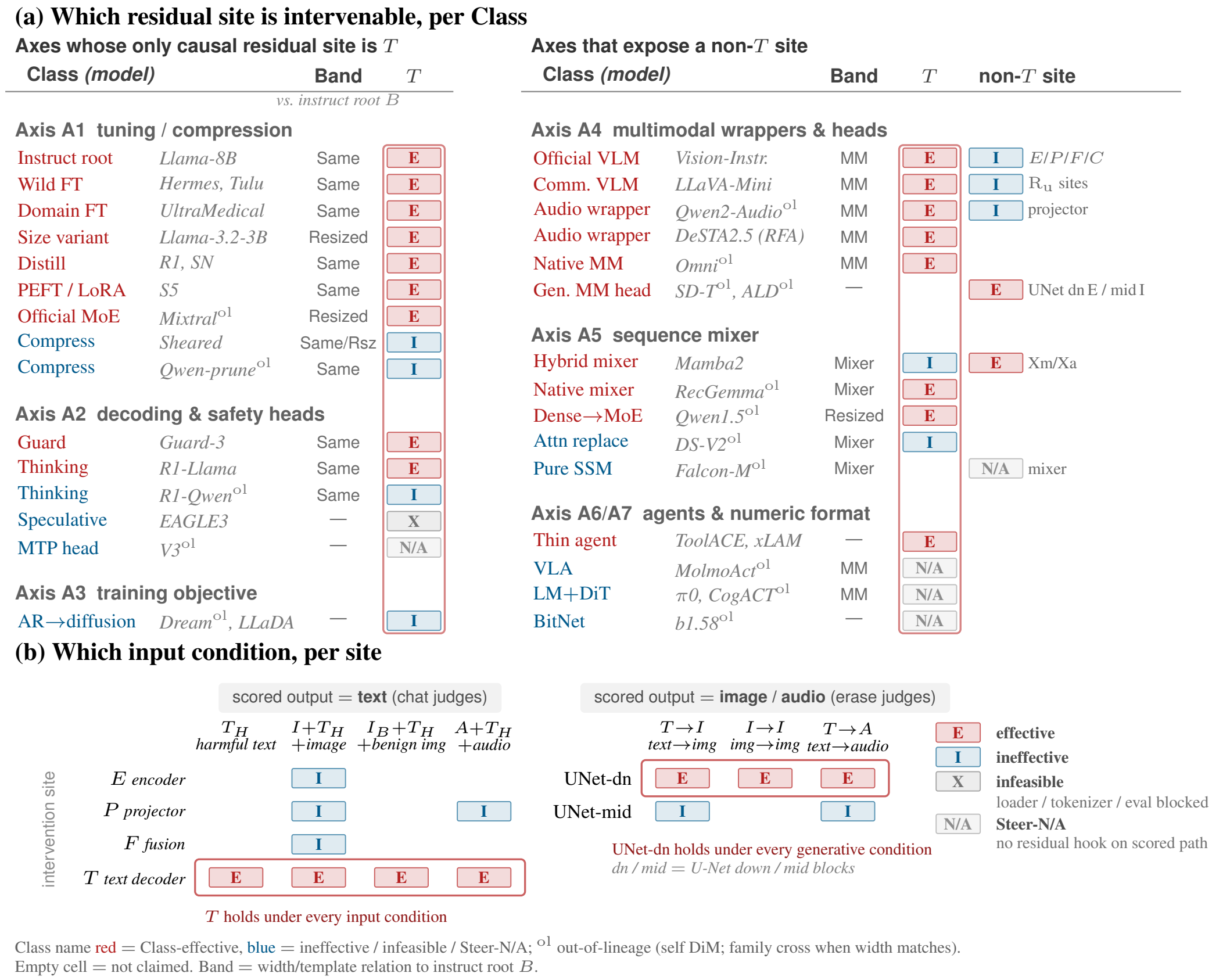

\paragraph{What each Axis changes.}
Axes mark architectural change relative to~$B$: weight-only tuning (A1),
decode / scoring-path change (A2), training-objective change (A3), attached
perception or generation stacks (A4), mixer replacement (A5), action / tool
readouts (A6), and numeric-format change (A7).
Whether each leaves a rank-1 residual site to intervene on is what
Figure~\ref{fig:class-site-grid} answers; detailed Axis prose is
App.~\ref{app:class-site-detail}.
Out-of-lineage\ool{} models without a matched donor report self DiM only.

\section{Text vs.\ non-text refusal directions (shared recipes)}
\label{sec:nt-dim}

We write the text-tower recipe used throughout Sec.~\ref{sec:emp}
(DiM $+$ rank-1 residual ablation at~$T$).
Non-text / conditioning DiM and optional RFA are introduced briefly below and
fully specified in App.~\ref{app:nt-dim-recipes}; Emp finds them ineffective
(or recipe-only) for headline chat attack credit.

\paragraph{Notation.}
Let $T$ denote the text-decoder residual stream and
$s\in\{E,P,F,C,Q,\mathrm{VT},\ldots\}$ index \emph{non-text} hook sites
($E$~$=$ perceptual encoder, $P$~$=$ modality projector,
$F$~$=$ fusion / cross-attention,
$C$~$=$ compressor or connector,
$Q$~$=$ Q-Former,
$\mathrm{VT}$~$=$ visual-/audio-token span \emph{inside}~$T$).
Write $u$ for a perceptual input (image~$x$, audio clip~$a$, \ldots),
$q$ for a fixed text probe, and $a_s(u;q)$ for the hooked activation at
site~$s$.
Let $\phi_s(\cdot)$ be a site-specific pool
(e.g.\ mean over patch tokens at~$E$, over projected image tokens at~$P$,
or last text position at~$T$).
Let $y$ be the scored model output: generated text for chat ASR
(Sec.~\ref{sec:emp}), or image/audio for generative output-modality probes.
We write $\mathrm{R}_{\ast}$ for the \emph{extraction contrast}
(Table~\ref{tab:dim-contrast}): what is averaged on each DiM side.
Hook site and Class may still differ under the same $\mathrm{R}_{\ast}$.

\paragraph{Text DiM $\mathrm{R}_{\mathrm{txt}}$ (condition~(a); primary attack).}
Following Arditi et~al.~\citep{arditi2024refusal},
contrasts are harmful vs.\ benign \emph{text} prompts
$\mathcal{D}_{\mathrm{harm}}^{\mathrm{txt}}$ /
$\mathcal{D}_{\mathrm{ben}}^{\mathrm{txt}}$
(no image, or image held fixed in a separate ablation):
\begin{align}
\mu_{\mathrm{txt}}^{(h)}
&=
\mathbb{E}_{p\sim\mathcal{D}_{\mathrm{harm}}^{\mathrm{txt}}}
\big[\phi_{T}\big(h^{(L)}(p)\big)\big],
\quad
\mu_{\mathrm{txt}}^{(b)}
=
\mathbb{E}_{p\sim\mathcal{D}_{\mathrm{ben}}^{\mathrm{txt}}}
\big[\phi_{T}\big(h^{(L)}(p)\big)\big],
\label{eq:mu-txt}
\\
\hat{v}_{\mathrm{txt}}
&=
\frac{\mu_{\mathrm{txt}}^{(h)}-\mu_{\mathrm{txt}}^{(b)}}
{\bigl\|\mu_{\mathrm{txt}}^{(h)}-\mu_{\mathrm{txt}}^{(b)}\bigr\|},
\quad
h \leftarrow h - \lambda\,(h^{\top}\hat{v}_{\mathrm{txt}})\,\hat{v}_{\mathrm{txt}}.
\label{eq:v-txt}
\end{align}
Guard label miss and hybrid Xm/Xa leaves use this \emph{same}
$\mathrm{R}_{\mathrm{txt}}$ family (different readout or block, not a new contrast).
Pure ablation in~\eqref{eq:v-txt} is the default intervention; optional RFA is
defined in App.~\ref{app:nt-dim-recipes}.

\paragraph{Non-text / conditioning DiM and RFA (brief).}
Two further DiM families appear in site audits but not in the headline ASR grid:
$\mathrm{R}_{\mathrm{u}}$ (harm vs.\ benign \emph{perceptual}~$u$ at non-text sites~$s$) and
Yin-style $\mathrm{R}_{\mathrm{cond}}$ (image$+$text vs.\ text-only conditioning at~$T$)
\citep{yin2026crossarch}.
Empirically both fail as chat attack extracts beyond text@$T$
(App.~\ref{app:nt-dim-recipes}).
Optional refusal-feature ablation (RFA) patches along the same~$\hat{v}$ instead of
zeroing~\citep{yu2025refat}; we treat it as a within-site recipe, not a Class
(\eqref{eq:rfa-bar}--\eqref{eq:rfa}; App.~\ref{app:nt-dim-recipes}).

\paragraph{Intervention: pure ablation (default).}
\label{para:rfa-general}
Write $a$ for the hooked activation at the extract$=$intervene site
($h$ at~$T$, or $a_s$ at a non-text site) and $\hat{v}$ for any unit DiM direction.
\textbf{Pure} rank-1 ablation~\citep{arditi2024refusal} removes the
component along~$\hat{v}$:
\begin{equation}
a \leftarrow a - \lambda\,(a^{\top}\hat{v})\,\hat{v}.
\label{eq:pure-abl}
\end{equation}
At $\lambda{=}1$ this zeroes the $\hat{v}$-coordinate; operating points here use
$\lambda{>}1$ (e.g.\ $4$ or $8$), which over-ablates / steers along the same axis.
RFA instead patches to the negative-side mean projection~$\bar{r}$
(\eqref{eq:rfa-bar}--\eqref{eq:rfa} in App.~\ref{app:nt-dim-recipes}).
Headline chat ASR uses~\eqref{eq:pure-abl}.

\paragraph{Structural contrast (families and sites).}
Table~\ref{tab:dim-recipe} catalogs \emph{extraction} families
($\mathrm{R}_{\mathrm{txt}}$ vs.\ $\mathrm{R}_{\mathrm{u}}$ hook/pool/path, plus all
extraction IDs); default intervene is rank-1 ablation in~\eqref{eq:v-txt}/
\eqref{eq:pure-abl} (optional RFA in App.~\ref{app:nt-dim-recipes}).
Table~\ref{apptab:site-class-map} maps Classes to those IDs
(so $\hat{v}_{\mathrm{txt}}$, Guard~$\hat{v}$@$T$, and hybrid Xm/Xa share
$\mathrm{R}_{\mathrm{txt}}$).

\begin{table}[t]
\centering
\caption{DiM \emph{extraction} families and recipes (not an intervene operator table).
Top block: $\mathrm{R}_{\mathrm{txt}}$ vs.\ $\mathrm{R}_{\mathrm{u}}$ (hook/pool/path to scored $y$).
Bottom: extraction IDs (same ID $=$ same pooling contrast; hook site and Class may still differ).
Intervene $=$ rank-1 ablation of the extracted $\hat{v}$ at the same site
(\eqref{eq:v-txt},~\eqref{eq:pure-abl}); optional RFA in App.~\ref{app:nt-dim-recipes}.
Yin-style $\mathrm{R}_{\mathrm{cond}}$~\citep{yin2026crossarch} contrasts \emph{modality / conditioning}
(image$+$text vs.\ text-only), \emph{not} harm vs.\ benign labels.
$\phi_s$ $=$ pool at site~$s$ before forming $\hat{v}$.}
\label{tab:dim-recipe}
\label{tab:dim-contrast}
\scriptsize
\setlength{\tabcolsep}{2.6pt}%
\renewcommand{\arraystretch}{1.05}%
\begin{tabularx}{\linewidth}{@{}l
  >{\raggedright\arraybackslash}X
  >{\raggedright\arraybackslash}X@{}}
\toprule
\multicolumn{3}{@{}l}{\emph{(A) Recipe axes}} \\
Axis & $\mathrm{R}_{\mathrm{txt}}$: $\hat{v}_{\mathrm{txt}}$@$T$ &
   $\mathrm{R}_{\mathrm{u}}$: $\hat{v}_{\mathrm{nt}}^{(s)}$@$s$ \\
\midrule
Contrast variable &
  text prompt $p$ &
  perceptual input $u$ (image~$x$, audio~$a$, \ldots) \\
Contrast pools &
  $\mathcal{D}_{\mathrm{harm/ben}}^{\mathrm{txt}}$ &
  $\mathcal{D}_{\mathrm{harm/ben}}^{(u)}$ (matched-format benign) \\
Hooked tensor &
  decoder residual $h^{(L)}$ &
  $a_s(u;q)$: e.g.\ $a_E$/$a_P$/$a_F$/$a_C$/$a_Q$ \\
Pool $\phi_s$ &
  $\phi_T$: often \emph{last text position} &
  $\phi_s$: mean over \emph{non-text} tokens (patches, VT, Q-Former);
  at fusion use image/audio positions, not last text \\
Path to $y$ &
  $T \!\to\! y$ &
  $E\!\to\!P\!\to\!\cdots\!\to\!T\!\to\! y$ \\
Typical $y$ &
  generated chat text &
  same text $y$ (VLM/A+T); separate judges for generative erase \\
\midrule
\multicolumn{3}{@{}l}{\emph{(B) Extraction ID catalog}} \\
ID & Contrast pools (DiM sides) & Typical $\hat{v}$ / use \\
\midrule
$\mathrm{R}_{\mathrm{txt}}$ &
  harmful vs.\ benign \emph{text} prompts &
  $\hat{v}_{\mathrm{txt}}$@$T$ (primary attack) \\
$\mathrm{R}_{\mathrm{u}}$ &
  harmful vs.\ benign \emph{percept}~$u$ (fixed probe~$q$) &
  $\hat{v}_{\mathrm{nt}}^{(s)}$@$E$/$P$/$F$/$\ldots$
  (shared residual protocol) \\
$\mathrm{R}_{\mathrm{cond}}$ &
  \emph{image$+$text} vs.\ \emph{text-only}
  (same prompt; Yin ABL\_POS / $\Delta h$) &
  $\hat{v}_{\mathrm{cond}}$@$T$ (defence / extract audit) \\
$\mathrm{R}_{\mathrm{concept}}$ &
  concept vs.\ neutral generation prompts &
  $\hat{v}_{\mathrm{concept}}$@UNet (T$\to$I / T$\to$A erase) \\
ctrl &
  matched-norm random; $\mu_{\mathrm{harm}}$ alone; PC1; blank canvas &
  specificity / recipe ablations \\
map &
  $\mathrm{R}_{\mathrm{txt}}$ then linear $W$ (width change) &
  $W\hat{v}_B$ on Size variant \\
\bottomrule
\end{tabularx}
\end{table}

\paragraph{Controls to verify.}
(i)~Matched-format benign images (avoid canvas / OCR confounds).
(ii)~Random unit vector of matched norm at the same site
(ctrl in Table~\ref{tab:dim-recipe}).
(iii)~Pool over \emph{image} positions at fusion layers---not last text tokens.
(iv)~Empty / collapsed generation rate near zero before reading ASR.

\section{Experimental setup}
\label{sec:setup}
We evaluate where a shared rank-1 extract$=$intervene protocol still hooks across
derivative Classes (Fig.~\ref{fig:class-site-grid}a): a localization audit under
fixed fairness gates, not an architecture-universal transfer claim.
Three evaluation bands stay separate:
\textbf{(chat)} AdvBench / HarmBench / StrongREJECT with chat judges;
\textbf{(conditional)} mixed-input cells (HADES, spoken AdvBench);
\textbf{(non-text site)} $E$/$P$/$F$ / dual audits under the same protocol
(Sec.~\ref{sec:nt-dim}; evidence in Sec.~\ref{sec:emp}).

\textbf{Benchmark split.}
AdvBench~\citep{zou2023advbench} uses disjoint prefix/suffix splits for DiM
extract ($n{=}128$) vs.\ ASR eval ($n{=}392$); Alpaca~\citep{taori2023alpaca}
supplies benign contrast only.
HarmBench ($n{=}320$) and StrongREJECT ($n{=}313$) are eval-only and never enter
DiM extraction (Table~\ref{apptab:benchmarks}).
HADES~\citep{li2024hades} and AdvBench TTS appear in conditional cells only.

\textbf{Per-Class task and DiM.}
Each red Class maps to one scored I/O task (App.~\ref{app:task-coverage}):
text chat ($T_H\!\to\!T$), Guard labels, or conditional
$I_H{+}T_H\!\to\!T$ / $A_H{+}T\!\to\!T$
(subscripts H/B $=$ harmful / benign-or-neutral inputs at eval time;
e.g.\ $I_H{+}T_H$ $=$ harmful image paired with harmful text).
Chat-band wrappers report $\mathrm{R}_{\mathrm{txt}}$@$T$ on the same three text
suites so ASR is comparable; multimodal-only eval is not mixed into that matrix.
Shared controls: harm--benign DiM, extract$=$intervene, rank-1 ablation
(\eqref{eq:pure-abl},~\eqref{eq:v-txt}), optional RFA (\eqref{eq:rfa}),
greedy decoding, and matched-norm random at the same site
(Sec.~\ref{sec:nt-dim}).
DiM's negative-side mean $\mu^{(-)}$ and the RFA patch $\bar{r}$ are related but
not cancelling (Sec.~\ref{para:rfa-general}).
Frozen layer/$\lambda$ are fixed before final generation
(Table~\ref{apptab:eval-set-iclr}); pre-specified exceptions are listed in
App.~\ref{app:setup-notes}.

\textbf{Judging and Class credit.}
Class attack \statusEff{} needs Nemotron$\wedge$HarmBench-cls (HBC) above base,
matched-norm random near floor, and a quality gate (non-empty / non-collapsed);
keyword (KW) ASR alone never qualifies.
A red Class claim uses the same gates that make negatives meaningful: disjoint
AdvBench extract/eval, HB/SR never in DiM extract, matched-norm random near base
(not near self), and a frozen layer/$\lambda$.
Steer-N/A marks an undefined residual path; \statusIneff{} stays open to other
extractors until pure matched-norm controls fail.
Bench saturation (e.g.\ Hermes Adv/SR \statusEff{} but tiny HB headroom) is a
model caveat, not Class blue.
Plotted chat markers are seven-judge median with IQR bars
(Fig.~\ref{fig:headline-asr}; calibration Table~\ref{apptab:judge-calib}); the
panel is thickness, not a Class recolor.
Production chat targets AdvBench $n{=}392$, HarmBench $n{=}320$, StrongREJECT
$n{=}313$; mechanism / dual / generative / ceiling cells often use smaller~$n$.
Glyphs on Fig.~\ref{fig:headline-asr}
($\bullet$~self; $\oslash$~family donor; $\blacksquare$~label task;
$\triangle$~mixer site) tag protocol, not Class color
(App.~\ref{app:headline-glyphs}).

\textbf{Residual comparison panels.}
Supporting Class coverage and residual bake-offs appear after the primary
mechanism result in Sec.~\ref{sec:emp}
(Figs.~\ref{fig:headline-asr},~\ref{fig:method-compare}; AdvBench $n{=}392$;
seven-judge median$+$IQR; HADES dropped for suite parity in the bake-off):
(a)~Vision residual-recipe bake-off (text DiM@$T$ vs.\ ABL\_POS / CMRM-style /
ShiftDC-style); ABL\_POS is a Yin-style defence / conditioning extract scored as
an attack---near-floor ASR means it is not a useful attack recipe, not that
``images make ASR low'';
(b)~$K{=}6$ Class-leaf lift (base / self / cross / random);
(c)~pure vs.\ RFA.
Related protocol coordinates are App.~\ref{app:rw-map}.

\section{Empirical localization}
\label{sec:emp}
\begin{figure}[!t]
\centering
\includegraphics[width=0.98\linewidth]{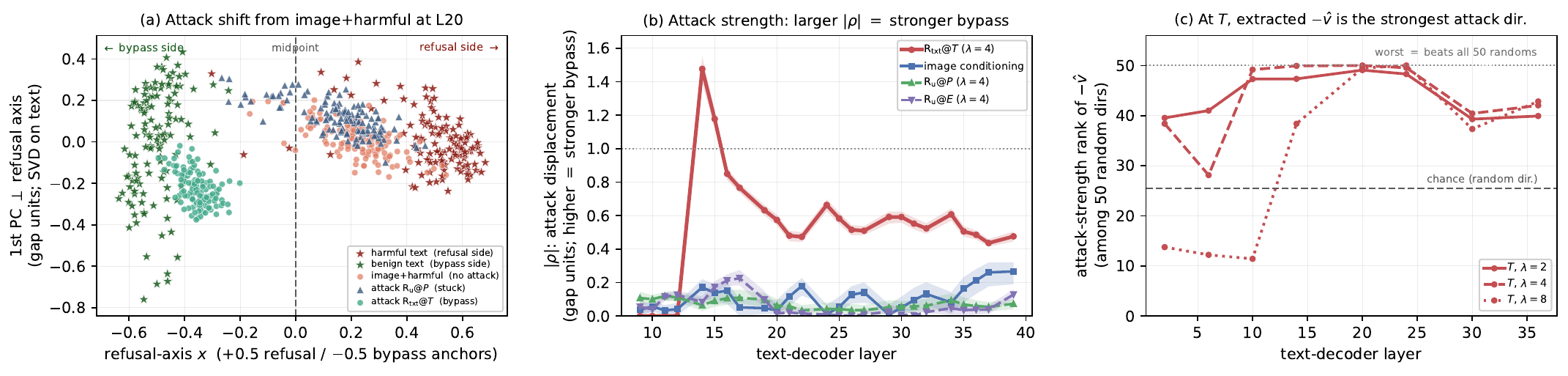}
\caption{Mechanism: intervenable lift concentrates at~$T$ under mean / rank-1 DiM
(Vision-Instruct, HADES; $n{=}128$).
The recipe is extremal at~$T$ and near chance at~$P$/$E$ among $50$ matched-norm
probes---a site asymmetry for this extractor, not a claim that no other feature
family could succeed.
(a)~Image$+$harmful inputs: $\mathrm{R}_{\mathrm{txt}}$@$T$ moves prompts to the bypass
side; $\mathrm{R}_{\mathrm{u}}$@$P$ stays on the refusal side
($x{=}{+}0.5$ refusal / ${-}0.5$ bypass; $y$ $=$ orthogonal PC).
(b)~Attack displacement $|\rho|$ ($99\%$ bootstrap): ${\approx}0.64$ at~$T$ vs.\
${\lesssim}0.12$ at $P$/$E$.
(c)~Rank of $-\hat{v}$ vs.\ $50$ matched-norm randoms on the safety margin
(rank~$50$ $=$ strongest; ${\approx}25.5$ $=$ chance). Mid-depth~$T$ is worst-case;
$P$/$E$ near chance.}
\label{fig:mechanism}
\end{figure}

\disc{We map where mean / rank-1 DiM refusal remains \emph{intervenable} under one
shared extract$=$intervene protocol, leading with the $T$ vs.\ non-text
asymmetry (Fig.~\ref{fig:mechanism}) under fixed gates; Class coverage and
residual bake-offs follow as supporting panels.}
Three claims line up with Fig.~\ref{fig:class-site-grid} and
Table~\ref{tab:localization}:
(1)~tested mixed-input cells (HADES; TTS) are \statusEff{} at~$T$, while mean
$\mathrm{R}_{\mathrm{u}}$ at $E$/$P$/$F$ is \statusIneff{};
(2)~Llama-lineage~$T$ is broadly \statusEff{}, with Thinking Class-effective only
under suite qualification; hard-Class probes in App.~\ref{app:fair-tables};
(3)~UNet-down concept erase is \statusEff{} while mid stays \statusIneff{}
(judges not pooled with chat).
Supporting patterns---audio projector \statusIneff{}, Thinking strong on
JailbreakBench (JBB) / HarmBench (HB) / StrongREJECT (SR) with AdvBench under
pure ablation still weak, RFA as a within-site variant---appear in the Class
maps; closed blue searches are in App.~\ref{app:fair-tables}.
Chat credit uses AdvBench extract$[0{:}128]$ / eval$[128{:}520]$ ($n{=}392$) with
HB/SR eval-only; judges stay separated by output type
(Nemotron$\wedge$HBC; NudeNet$+$CLIP; CLAP).

\paragraph{Visual-harm control (VLGuard).}
\disc{Because HADES pairs adversarial images with explicitly harmful text, we also
evaluate VLGuard Unsafe prompts ($n{=}200$) under a single Llama-Guard-3 success
definition (empty rate $0$).
At the paper Vision operating point (L10, $\lambda_t{=}4$, $\lambda_{\mathrm{mm}}{=}2$), ASR moves
\dischl{{$0.030{\to}0.150$}} at~$T$ ($\Delta{+}0.120$) versus $+0.020$ at~$P$ and $+0.035$ at~$E$
(random stays near base).
We treat this as a localization-\emph{direction} control against a pure
text-payload artifact, not as a HADES-scale absolute attack win; keyword ASR is
not used for the VLGuard claim.
As a pooling fairness check at the same $\lambda_{\mathrm{mm}}{=}2$ OP, replacing mean
patch pooling by max pooling for @$E$ DiM extract does \emph{not} increase LG
lift (mean \dischl{{$\Delta{+}0.030$}}; max \dischl{{$\Delta{-}0.010$}}; $n{=}200$), while extract-time
$|a^{\top}\hat{v}|$ is larger under max than mean and both dwarf matched-norm
random---so flat @$E$ is not explained by mean dilution alone.
Scale sensitivity is quality-gated: $\lambda_{\mathrm{mm}}{=}8$ yields mostly
collapsed generations and is discarded; at $\lambda_{\mathrm{mm}}{=}4$ mean-pool
$n{=}200$, coherent LG rises \dischl{{$0.035{\to}0.232$}} ($\Delta{+}0.197$, gibber $11.5\%$)
while max stays flat ($\Delta{-}0.005$), so we do \emph{not} claim @$E$ is inert
at every multiplier---only that under the paper OP (and max-pool at that OP)
lift remains small relative to @$T$.
Judge-threshold sensitivity on the primary pack keeps $T\Delta{>}E\Delta$ under
LG / Nemo / HBC / Nemo$\wedge$HBC / majority-2.
Within-$E$ matched edit budgets (same $\lambda|a^{\top}\hat{v}|$ for DiM@$\lambda{=}2$
and scaled random) leave DiM LG near the paper OP
(\dischl{{$0.035{\to}0.075$}}) while matched-budget random rises to $0.32$---equalizing
edit size does not turn mean DiM@$E$ into a strong attack.}

\textbf{Site asymmetry under the tested mean / rank-1 recipe.}
\label{sec:mech}
A flat non-text cell admits three readings that matched-norm random alone cannot
separate: the refusal-related axis may be absent, present while our mean /
rank-1 recipe misses it, or found while~$T$ re-imposes refusal downstream.
If the weak-recipe reading is true for our extractor, every non-text
\statusIneff{} cell is an artefact of that extractor rather than a property of
the site.
Fig.~\ref{fig:mechanism} does not exclude every alternative extractor
(multi-direction or SAE features remain open), but on the Vision-Instruct leaf
that carries the HADES win, the same mean / rank-1 protocol that is extremal
at~$T$ is near chance at~$P$/$E$.

\begin{table}[!t]
  \centering
\caption{Emp claim ledger under one extract$=$intervene protocol.
Complements Fig.~\ref{fig:class-site-grid}: one row $=$ one localization claim
(mixed@$T$ / $\mathrm{R}_{\mathrm{u}}$ flat / catalog / UNet; dual supporting).
Primary $=$ Nemotron ASR $0\!\to\!$steer (erase $=$ NudeNet / CLAP; never pooled with chat).
Control $=$ matched-norm random, suite headroom, or generative mid-block null.
Dual rows: $\Delta$ $=$ ASR lift (steer$-$base); slash $=$ $T$-alone / dual on the
same pack ($\neq$ HADES test500).
Class ASR: Fig.~\ref{fig:headline-asr}; I/O$\times$judge: App.~\ref{app:task-coverage}.}
  \label{tab:localization}
  \scriptsize
  \setlength{\tabcolsep}{2.2pt}%
  \renewcommand{\arraystretch}{1.08}%
  \begin{tabularx}{\linewidth}{@{}
    l
    >{\raggedright\arraybackslash}X
    c
    >{\raggedright\arraybackslash}X
    >{\raggedright\arraybackslash}X
  @{}}
  \toprule
  Claim & Evidence cell & Site & $0\!\to$steer (Nemo) & Control / note \\
  \midrule
  \multicolumn{5}{@{}l}{\emph{(1) Coordinates: mixed-input at~$T$; non-text flat}} \\
  Mixed@$T$ & Vision HADES test500 & $T$ & $0.034\!\to\!0.616$ & rand $0.024$ \\
  Mixed@$T$ & Qwen2-Audio TTS Adv & $T$ & $0.03\!\to\!0.84$ & held-out Adv text \\
  $\mathrm{R}_{\mathrm{u}}$ flat & Vision @$P$/$F$ & $P$/$F$ & $0.021\!\to\!0.021$ & $=$ matched-norm rand \\
  $\mathrm{R}_{\mathrm{u}}$ flat & Qwen2-Audio @$P$ & $P$ & $0.02\!\to\!0.03$ & benign-TTS contrast \\
  ABL\_POS & LLaVA $\mathrm{R}_{\mathrm{cond}}$@$T$ & $T$ & $0.38\!\to\!0.24$ & defence role; Tab.~\ref{apptab:asr-bakeoff} \\
  \addlinespace
  \multicolumn{5}{@{}l}{\emph{(supporting) Dual ablations (pack $\neq$ HADES test500)}} \\
  Dual$\approx T$ & Vision $T{+}P$ & $T{+}P$ & $\Delta$ $0.19$ / $0.21$ & lift: $T$ alone / dual \\
  Dual$\approx T$ & LLaVA-Mini $T{+}C$ & $T{+}C$ & $\Delta$ $0.35$ / $0.40$ & lift: $T$ alone / dual; $C$ alone flat \\
  \addlinespace
  \multicolumn{5}{@{}l}{\emph{(2) Catalog breadth under audited controls}} \\
  Llama~$T$ & Hermes Adv392 & $T$ & $0.23\!\to\!0.69$ & HB headroom tiny ($0.38\!\to\!0.43$) \\
  Llama~$T$ & ToolACE Adv392 & $T$ & $0\!\to\!0.82$ & HBC $0\!\to\!0.73$; rand$\approx$base \\
  \ool~MoE & Mixtral Adv392 & $T$ & $0.03\!\to\!0.86$ & HBC $0.04\!\to\!0.82$; rand$\approx$base \\
  Native MM & Omni Adv392 & $T$ & $0.04\!\to\!0.66$ & EB-family cross $0.81$ \\
  Native MM & Omni HB / SR & $T$ & $0.09\!\to\!0.40$ / $0.01\!\to\!0.58$ & same OP; rand$\approx$base \\
  Hard Classes & App.~\ref{app:fair-tables} & $T$ & Audio@$T$ unlock & mixed measured / N/A / not-run \\
  \addlinespace
  \multicolumn{5}{@{}l}{\emph{(3) Second band (generative; judges not pooled with chat)}} \\
  Erase & SD-Turbo T$\to$I & UNet-dn & $0.79\!\to\!0$ & mid-block \statusIneff{} \\
  Erase & SD-Turbo I$\to$I & UNet-dn1 & $1.0\!\to\!0$ @$\lambda{=}25$ & down2 / mid flat \\
  Erase & AudioLDM T$\to$A & UNet-dn & erase (CLAP) & mid-block \statusIneff{} \\
  \bottomrule
  \end{tabularx}
\end{table}

\begin{figure}[p]
  \centering
  \includegraphics[width=\linewidth]{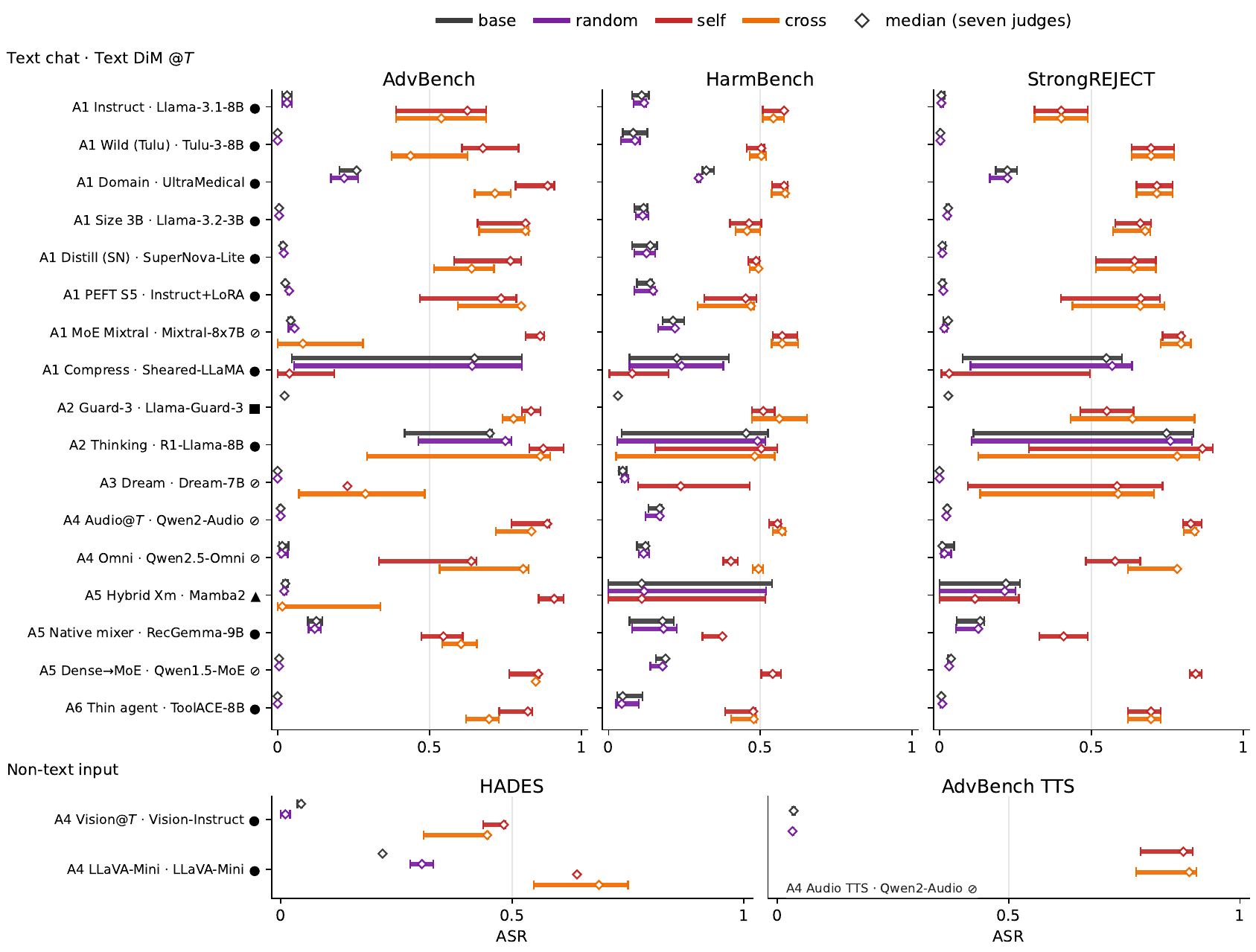}
  \caption{Class-leaf bypass ASR under Text DiM @$T$ (\eqref{eq:pure-abl}).
  Top: AdvBench $n{=}392$ / HarmBench $n{=}320$ / StrongREJECT $n{=}313$;
  bottom: HADES and AdvBench TTS (suite-split).
  Per leaf: base / random / self / cross; diamond $=$ seven-judge median; thin bars $=$ IQR
  (Class gates Nemotron$\wedge$HarmBench-CLS).
  Y-axis glyphs: App.~\ref{app:headline-glyphs}; OPs: Table~\ref{apptab:eval-set-iclr}.}
  \label{fig:headline-asr}
  \end{figure}

  \begin{figure}[p]
  \centering
  \includegraphics[width=\linewidth]{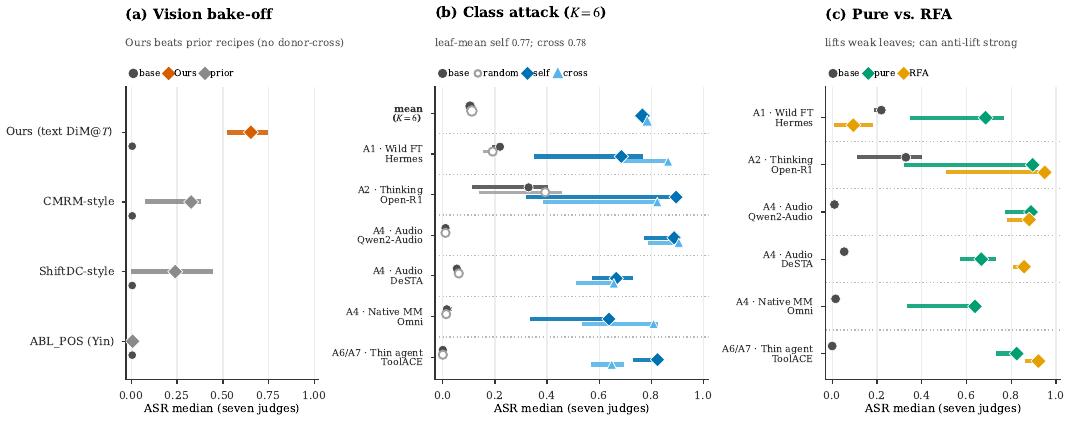}
  \caption{Three residual-attack claims on AdvBench $n{=}392$
  (not Class coverage---Fig.~\ref{fig:headline-asr}).
  Median ASR$+$IQR as in Fig.~\ref{fig:headline-asr}.
  (a)~Vision recipe bake-off (text DiM@$T$ vs.\ CMRM-style / ShiftDC-style / ABL\_POS).
  (b)~$K{=}6$ leaf attack lift (base / random / self / cross).
  (c)~Pure vs.\ RFA@$T$.}
  \label{fig:method-compare}
\end{figure}

We cache per-layer text-stream residuals once ($n{=}128$ HADES
dev for extraction, disjoint test for evaluation) and score three quantities on
the same cache, relative to the harm$\to$benign contrast~\citep{yu2025refat}.

\textbf{Where an intervention moves a prompt.}
Writing $\hat{v}_{\mathrm{txt}}$ for the unit text DiM and $g$ for the
harm$\to$benign gap norm, prompt position
$x = \langle h - c,\ \hat{v}_{\mathrm{txt}}\rangle / g$ places harmful text at
$x{=}{+}0.5$ and benign at $x{=}{-}0.5$ by construction ($c$ midpoint).
At $\lambda{=}4$, $\mathrm{R}_{\mathrm{u}}$@$P$/$E$ carry only $13$--$15\%$ of
image$+$harmful prompts past the midpoint (vs.\ $5\%$ unattacked), while text
ablation at the same $\lambda$ carries $100\%$ (Fig.~\ref{fig:mechanism}a).

\textbf{How far along the axis it moves.}
The projection ratio $\rho = \langle \Delta h,\ \hat{v}_{\mathrm{txt}}\rangle / g$
measures shift in gap units (avoiding raw-cosine inflation).
Downstream of its intervention layer, text ablation gives $|\rho|{=}0.64$; image
conditioning and $\mathrm{R}_{\mathrm{u}}$@$P$/$E$ stay at $0.07$--$0.11$ with
$99\%$ bootstrap bands that never approach the text curve
(Fig.~\ref{fig:mechanism}b).

\textbf{Whether the direction is the strongest matched-norm attack.}
On a teacher-forced safety margin (refusal vs.\ complying continuation), we
apply~\eqref{eq:pure-abl} along the attack-signed direction $-\hat{v}$
(the sign cancels in~\eqref{eq:pure-abl}; kept for ranking polarity) and compare
against $50$ matched-norm randoms~$u_i$.
Rank $50/50$ means our direction is the strongest probe attack; ${\approx}25.5$
is chance.
\emph{Leverage} is
$|\Delta\mathrm{margin}(-\hat{v})| /
\mathrm{mean}_i|\Delta\mathrm{margin}(u_i)|$
(${\approx}1$ $=$ no stronger than chance).
At~$T$, $-\hat{v}$ ranks $50/50$ at mid depth ($\lambda{=}4$; leverage $6.0$),
while $P$/$E$ return rank $28$--$32$ and leverage $\le 0.26$
(Fig.~\ref{fig:mechanism}c).
These results pressure a weak-recipe reading for this extractor at perception
sites without separating absence from downstream reimposition, and without
exhausting multi-direction search.
The worst-case property at~$T$ holds in a mid-depth, moderate-$\lambda$ window
(at $\lambda{=}8$ early layers, randoms can match the margin drop).

\textbf{The same instrument screens donors.}
Mapper-free cross asks whether a donor $\hat{v}_B$ approximates the recipient's
refusal direction.
Template-matched subspace overlap $R$ against a matched-norm random band is a
necessary screen on the pairs we tested: donors inside the band transfer
nothing, and every donor above the band transfers
(Table~\ref{apptab:donor-subspace}); Mixtral is the clean within-recipient swap
(Llama donor flat; Mistral-family donor $R{=}0.63$, ASR$\!\to\!0.98$).
Overlap is not sufficient (e.g.\ Tulu recovers only a fraction of self lift), so
usable-donor screening remains separate from Class / site effectiveness.

\paragraph{Non-text / conditioning DiM.}
Mean non-text DiM at $E$/$P$/$F$/VT and Yin-style $\mathrm{R}_{\mathrm{cond}}$ do not
yield stable chat attack \statusEff{} beyond text@$T$; dual ablations are
pack-noted supporting controls (App.~\ref{app:nt-dim-recipes}).

\paragraph{Text tower~$T$.}
Text DiM with rank-1 ablation at~$T$ raises ASR on held-out AdvBench and on
mixed-input cells (HADES $I_H{+}T_H$; spoken AdvBench) while matched-norm
random stays near base; within-family cross holds when widths match
(Fig.~\ref{fig:headline-asr}).
Cross-condition $I_B{+}T_H$ tracks harmful text, not image content.
Thinking is Class-effective on JBB/HB/SR (and Distill Adv392) while AdvBench
under pure ablation remains weak.
On a shared HADES leaf, Vision-Instruct@$T$ is the clean attack win above random;
Yin ABL\_POS lowers with-image ASR as a defence
(Table~\ref{apptab:asr-bakeoff}; Fig.~\ref{fig:method-compare}a).
FigStep does not reproduce the HADES win and caps ``mixed-input'' generalization.

\paragraph{Guard and audio wrappers.}
Llama-Guard-3 raises harmful$\to$safe miss $0.03\!\to\!0.88$ under the same
$\hat{v}_{\mathrm{txt}}$@$T$ protocol (label miss, not chat ASR).
Qwen2-Audio is Class \statusEff{} at~$T$ (including TTS/silence), while
projector $\mathrm{R}_{\mathrm{u}}$ stays \statusIneff{}.
RFA is a within-site recipe, not a new Class: it can lift weak-pure leaves
(DeSTA; ToolACE; Qwen2-Audio) and hurt already-strong ones
(Hermes pure ${\approx}0.69$ vs.\ RFA median ${\approx}0.10$;
Fig.~\ref{fig:method-compare}c).

\paragraph{Yin $\mathrm{R}_{\mathrm{cond}}$ and generative erase.}
Ablating Yin-style $\hat{v}_{\mathrm{cond}}$ acts as a defence / extract audit at~$T$,
not as an inverse of $\mathrm{R}_{\mathrm{txt}}$ (App.~\ref{app:nt-dim-recipes}).
Concept DiM in SD-Turbo / AudioLDM UNets erases harm at down sites
(T$\to$I NudeNet $0.79\!\to\!0$; T$\to$A CLAP erase) and fails at mid under the
same judges; Dream / LLaDA remain Steer-N/A or empty-gated \statusIneff{}.

\paragraph{Reading Fig.~\ref{fig:class-site-grid}a.}
\textcolor{evalcls}{Red} marks gated \statusEff{} (including suite-qualified
Thinking); \textcolor{outcls}{blue} marks \statusIneff{}, \statusInfeas{}, or
\steerNA{}.
Catalog-lineage, out-of-lineage\ool{}, and hard-Class probes are separate credit
sources (App.~\ref{app:fair-tables}).

\textbf{Scope, limits, and outlook.}
\label{sec:limits}
$B\!\to\!M$ cross is primarily Llama-3.1-8B-Instruct derivatives; \ool{} leaves
fill Axis gaps with width-matched family donors when needed
(App.~\ref{app:fair-tables} / \ref{app:limits-detail}; Related
App.~\ref{app:rw-map}).
Production chat claims use AdvBench / HarmBench / StrongREJECT at the~$n$ in
Fig.~\ref{fig:headline-asr}; mechanism / dual / generative / ceiling cells often
use smaller~$n$.
Thinking is suite-qualified red, not an unqualified Class win; KW ceilings
(FigStep) do not relabel Classes; Guard is label miss; generative erase uses
separate judges.
RFA and defence-style extracts are within-site recipes
(Fig.~\ref{fig:method-compare}).
For \statusIneff{}, \statusInfeas{}, and \steerNA{} cells, searching other
refusal-steering families (multi-direction / SAE / non-residual protocols) is
left as future work---this paper maps where the shared mean / rank-1 DiM still
hooks.

\section{Conclusion}
\label{sec:conclusion}
Safety-relevant behavior in open-weight multimodal derivatives is often audited
as if a single residual hook were enough.
\disc{Under one shared mean / rank-1 difference-in-means (DiM) protocol with
matched-norm random controls, the chat attack lift we measure is
\emph{intervenable} primarily at the text decoder~$T$: HADES image$+$text and
spoken AdvBench raise attack success rate at~$T$, while the same mean recipe at
encoders, projectors, and fusion ($E$/$P$/$F$) does not.}
\disc{This is a claim about intervention accessibility under this DiM family---not
that refusal features are absent outside~$T$, and not a fully site-isolated
effect when contrasts and pooling change with the hook.}
Margin-rank and leverage instruments pressure a weak-extractor reading of those
flat perceptual cells for this DiM family; alternatives on ineffective,
infeasible, and N/A leaves are future work.
\disc{VLGuard Unsafe ($n{=}200$; Llama-Guard) preserves $T{>}P$/$E$ directionality at
modest absolute ASR under the paper OP, with pooling / scale / threshold /
matched-edit diagnostics reported in Sec.~\ref{sec:emp}.}
Catalog breadth, mapper-free $B\!\to\!M$ cross when widths match, and generative
U-Net erase under separate judges remain supporting coverage.
\disc{The primary result is the recipe-scoped $T$ vs.\ non-text asymmetry; the
Class$\times$site ledger records where this protocol still hooks under the same
gates.}

\section*{Ethics Statement}
This work evaluates gray-box refusal bypass on open-weight models for scientific
auditing.
We do not release ready-to-weaponize tooling beyond methods already public in the
refusal-direction literature.
Dual-use concerns around residual refusal attacks motivate mitigations studied
elsewhere~\cite{yu2025refat}.

\section*{Reproducibility Statement}
We report extract/eval splits, frozen layer/$\lambda$ operating points
(Table~\ref{apptab:eval-set-iclr}), matched-norm random / empty-rate gates, and the
seven-judge panel protocol used for production ASR ranges.
An anonymized archive with the core extract / intervene / matched-norm-random
scripts, frozen operating-point configs, and a minimal end-to-end example is
included in the supplementary material; the archive is a keyword-ASR smoke path,
while production Class credit uses Nemotron$\wedge$HBC as in the main text.
Full experiment scripts, panel JSON, and extended Class grids will be released
upon acceptance.

\section*{AI use statement}
Large language models assisted with grammar and copy-editing, and with drafting
and tightening of LaTeX prose.
Models did not run experiments, choose operating points, or invent numerical
results.
All technical claims, experimental design, and final wording are the authors'
responsibility.

\bibliography{iclr2027_conference}
\bibliographystyle{iclr2027_conference}

\newpage
\appendix
\FloatBarrier
\section{Benchmark splits and frozen operating points}
\label{app:setup}

\subsection{Setup protocol notes (OP exceptions and fairness reminders)}
\label{app:setup-notes}
Frozen layer/$\lambda$ per model are fixed before final generation
(Table~\ref{apptab:eval-set-iclr}).
Pre-specified exceptions (not suite-tuned) include Hermes/Tulu $\lambda{=}6$,
3B ridge $W$, Vision L10/L20 @$\lambda{=}4$, Omni mid-depth L14, LLaVA last-$k$
cross, Guard early cross, and R1 post-think scoring.
Existence / mechanism / generative-erase / ceiling cells may use fair Adv
$n{=}128$ on the disjoint suffix ($[392{:}520]$) or smaller~$n$ with task-native
judges---not pooled with production chat ASR.
Blue exploration grids and extended pure-vs-RFA ledgers are
App.~\ref{app:fair-tables}.

\subsection{Headline $y$-axis glyphs}
\label{app:headline-glyphs}
Fig.~\ref{fig:headline-asr} appends one glyph to each leaf name so the $y$-axis
states \emph{which} extract$=$intervene protocol the plotted ASR uses.
The markers are orthogonal to red/blue Class credit
(Fig.~\ref{fig:class-site-grid}a):
\begin{itemize}
\item $\bullet$~\textbf{self DiM.} Harm--benign Text DiM on the recipient;
intervene with the same $\hat{v}$ (most chat leaves; Root Instruct, ToolACE,
Audio@$T$, Omni, \ldots).
\item $\oslash$~\textbf{family donor.} Width-matched within-family $\hat{v}$ when
Llama $B\!\to\!M$ fails the random-band / transfer gates
(e.g.\ Mixtral$\leftarrow$Mistral; selected Qwen-family leaves).
\item $\blacksquare$~\textbf{label task.} Scored as Guard label-miss (keyword /
task-native), not as chat seven-judge ASR---do not pool with Adv/HB/SR diamonds.
\item $\triangle$~\textbf{mixer site.} Intervention at hybrid / native mixer
residuals (Mamba2 Xm/Xa multi-site) rather than a single text-tower depth.
\end{itemize}
A hollow $\circ$ marker is reserved for pure Llama $B\!\to\!M$ donor runs; the
current headline grid does not plot it as a separate $y$-tag (cross is already
the fourth diamond per leaf). Glyphs never override the Nemotron$\wedge$HBC
Class gate.

\begin{table}[H]
\centering
\caption{Extract vs.\ evaluation split (chat-band text suites).
HADES / TTS are conditional-input cells (not chat-band extract pools).
HarmBench, StrongREJECT, HADES, and TTS never enter DiM extraction.}
\label{apptab:benchmarks}
\footnotesize
\setlength{\tabcolsep}{3pt}%
\begin{tabularx}{\columnwidth}{@{}l l r r r >{\raggedright\arraybackslash}X@{}}
\toprule
Corpus & Mod. & Pool & Extract & Eval & Role \\
\midrule
AdvBench & text & 520 & 128 & 392 &
 Harmful requests; disjoint $[0{:}128]$ / $[128{:}520]$ \\
Alpaca & text & 31k & 128 & --- &
 Benign contrast for $\mathrm{R}_{\mathrm{txt}}$ only \\
HarmBench & text & 320 & --- & 320 & Eval-only \\
StrongREJECT & text & 313 & --- & 313 & Eval-only \\
HADES (step~5) & image & 750 & --- & 500 & conditional-input \\
AdvBench TTS & audio & 520 & --- & 392 & conditional-input \\
\bottomrule
\end{tabularx}
\end{table}

\begin{table}[H]
\centering
\caption{Chat-band operating points for Fig.~\ref{fig:headline-asr}.
\textbf{Extract:} Text DiM (harm--benign mean difference) at the listed residual site;
intervene $=$ rank-1 ablation of the same $\hat{v}$ (\eqref{eq:pure-abl}), not RFA
unless noted. $\hat{v}_B=$ Instruct (or family) donor; Size uses ridge $W\hat{v}_B$.
Eval $n$: AdvBench 392 / HB 320 / SR 313 (Table~\ref{apptab:benchmarks}); HADES 500;
TTS Adv 392. DeSTA RFA is Fig.~\ref{fig:method-compare} only.}
\label{apptab:eval-set-iclr}
\scriptsize
\setlength{\tabcolsep}{2pt}%
\renewcommand{\arraystretch}{1.05}%
\begin{tabularx}{\columnwidth}{@{}c
 >{\raggedright\arraybackslash}X
 l
 >{\raggedright\arraybackslash}X
 >{\raggedright\arraybackslash}X@{}}
\toprule
Ax. & Class (model) & Extract & Self OP & Cross OP \\
\midrule
A1 & Root (Llama-3.1-8B-Instruct)
 & Text DiM @$T$ & L16 $\lambda{=}2$ & same $\hat{v}$ \\
A1 & Wild FT (Hermes / Tulu-3 / UltraMedical)
 & Text DiM @$T$ & L16 $\lambda{=}6$ / $6$ / $4$ & $\hat{v}_B$ / self-donor / $\hat{v}_B$ \\
A1 & Size / Distill (3B / SuperNova)
 & Text DiM @$T$ & L14 $\lambda{=}4$ / L16 $\lambda{=}4$ & $W\hat{v}_B$ / $\hat{v}_B$ \\
A1 & PEFT S5 (Instruct$+$LoRA)
 & Text DiM @$T$ & L16 $\lambda{=}6$ & $\hat{v}_B$ \\
A1 & MoE Mixtral-8x7B
 & Text DiM @$T$ & L16 $\lambda{=}1$ & Mistral-family $\hat{v}$ \\
A1 & Compress (Sheared-LLaMA)
 & Text DiM @$T$ & L16 $\lambda{=}1$ (ineff) & --- / $\hat{v}_B$ \\
A2 & Guard-3
 & Text DiM @$T$ (KW miss) & L16 $\lambda{=}3.75$ & L12 $\lambda{=}3.75$, $\hat{v}_B$ \\
A2 & Thinking (R1-Llama)
 & Text DiM @$T$ & L16 $\lambda{=}6$ & $\hat{v}_B$ \\
A3 & Dream-7B
 & Text DiM @$T$ (DLM) & L14 $\lambda{=}2$ (fair); mid/multi sweep & family $\hat{v}$ \\
A4 & Vision-Instruct / LLaVA-Mini
 & Text DiM @$T$ on HADES & L10 $\lambda{=}4$ / L16 $\lambda{=}4$ & L20 $\lambda{=}4$ / $\hat{v}_B^{k}$ \\
A4 & Qwen2-Audio @$T$ / TTS
 & Text DiM @$T$ & L16 $\lambda{=}2$ & Qwen1.5-family $\hat{v}$ \\
A4 & Qwen2.5-Omni
 & Text DiM @$T$ & L14 $\lambda{=}2$ & EB-family $\hat{v}$ \\
A5 & Mamba2 (Xm/Xa)
 & Text DiM @mixer & Xm$\{14,16,18\}$/Xa$\{15,17,19\}$ $\lambda{=}3$ & $\hat{v}_B$ \\
A5 & RecGemma-9B / Qwen1.5-MoE
 & Text DiM @$T$ & L18 $\lambda{\approx}1.45$ / L12 $\lambda{=}2$ & $\hat{v}_B$ / family \\
A6 & ToolACE-8B (thin agent)
 & Text DiM @$T$ & L16 $\lambda{=}4$ & $\hat{v}_B$ \\
\bottomrule
\end{tabularx}
\end{table}

\begin{table}[H]
\centering
\caption{Judge calibration on gold160 held-out labels.
Plotted chat ASR uses the \emph{median} over the seven-judge panel
(keyword excluded) with IQR bars (Setup).}
\label{apptab:judge-calib}
\footnotesize
\setlength{\tabcolsep}{3pt}%
\begin{tabular}{@{}lrrrr@{}}
\toprule
Judge & Acc & Prec & Rec & Note \\
\midrule
Keyword (ref.) & 0.83 & 0.73 & 1.00 & excluded from median/IQR \\
Llama-Guard & 0.91 & 0.86 & 0.94 & panel \\
Nemotron SG 8B v3 & 0.91 & 0.90 & 0.89 & panel \\
MD-Judge-v0.1 & 0.91 & 0.84 & 0.97 & panel \\
BeaverDam-7B & 0.89 & 0.84 & 0.94 & panel \\
Open-weight LLM judge (26B) & 0.89 & 0.92 & 0.83 & panel \\
HarmBench-cls & 0.89 & 0.90 & 0.85 & HB suite \\
StrongREJECT @0.2 & 0.86 & 0.81 & 0.92 & SR suite \\
\bottomrule
\end{tabular}
\end{table}
\FloatBarrier

\section{Related-work protocol coordinates}
\label{app:rw-map}
Protocol-axis comparison vs.\ related settings (Role / Extract / Intervene /
matched-norm Random) is Table~\ref{tab:related-compare}.
Extract$=$intervene is enforced in our attack audits; ASTRA steers at~$T$ even
when attribution is visual (consistent with our $E$/$P$ \statusIneff{}---no ASTRA ASR row).
Same-pack Vision numbers: Fig.~\ref{fig:method-compare}a
(LLaVA footnotes: Table~\ref{apptab:asr-bakeoff}); localization: Table~\ref{tab:localization}.

\begin{table}[H]
\centering
\caption{Related settings vs.\ this work (protocol axes only).
Narrative context: Sec.~\ref{sec:rw}.
Same-pack numbers: Fig.~\ref{fig:method-compare}a /
Table~\ref{apptab:asr-bakeoff}; localization: Table~\ref{tab:localization}.}
\label{tab:related-compare}
\small
\setlength{\tabcolsep}{2.5pt}%
\begin{tabular}{@{}lcccc@{}}
\toprule
Approach & Role & Extract & Intervene & Random \\
\midrule
Arditi~\citep{arditi2024refusal} & attack & text DiM & $T$ & (add) \\
CMRM~\citep{liu2025cmrm} & defence & MM vs text & $T$ multi & --- \\
ShiftDC~\citep{zou2025shiftdc} & defence & modality shift & $T$ & --- \\
Yin ABL\_POS~\citep{yin2026crossarch} & defence & cond@$T$ & $T$ & yes \\
ASTRA~\citep{wang2025astra} & defence & visual attr & \textbf{$T$} & --- \\
ECSO~\citep{gou2024ecso} & defence & --- (I$\to$T) & pipeline & --- \\
\textbf{Ours} $\mathrm{R}_{\mathrm{txt}}$ & attack & text DiM & \textbf{same $T$} & \textbf{yes} \\
\textbf{Ours} $\mathrm{R}_{\mathrm{u}}$ & attack & harm image/audio & \textbf{same $E$/$P$/$F$} & \textbf{yes} \\
\textbf{Ours} concept & erase & concept DiM & \textbf{same UNet} & \textbf{yes} \\
\bottomrule
\end{tabular}
\end{table}
\FloatBarrier

\section{Non-text DiM, conditioning DiM, and RFA recipes}
\label{app:nt-dim-recipes}
Main-text Sec.~\ref{sec:nt-dim} keeps Text DiM @$T$ and pure ablation as the
primary attack recipe.
Full definitions of $\mathrm{R}_{\mathrm{u}}$, $\mathrm{R}_{\mathrm{cond}}$, and RFA,
plus the negative Emp audits, are collected here.

\paragraph{Where the hooks sit (Vision-style stack).}
\label{app:hook-diagram}
Across chat Axes/Classes that still expose an autoregressive text decoder, the
\emph{text} extract is structurally the same object: last-token (or last-$k$)
residual $h^{(\ell)}$ inside transformer block~$\ell$, DiM $=$ harm$-$benign mean
difference, intervene $=$ rank-1 ablation of that $\hat{v}$
(\eqref{eq:mu-txt}--\eqref{eq:v-txt}).
What changes by Class is the \emph{surrounding} stack (chat template, LoRA, MoE
routing, thinker prefix, mixer Xm/Xa), not the DiM algebra at~$T$.
Non-text sites attach \emph{before} or \emph{beside} that decoder: encoder tokens
($E$), projector / connector outputs ($P$), fusion / cross-attn ($F$), or
visual-token spans inside~$T$ ($\mathrm{VT}$).
Fig.~\ref{appfig:hook-sites} sketches the Vision-Instruct path; Audio / Omni
replace the image tower by an audio (or thinker) tower but keep the same
``pool at~$s$, DiM, ablate at~$s$'' template (\eqref{eq:mu-nt}--\eqref{eq:v-nt}).

\begin{figure}[!ht]
\centering
\begin{tikzpicture}[
  font=\footnotesize,
  >=Latex,
  x=2.05cm, y=1.00cm,
  box/.style={
    draw=black!35, line width=0.45pt, rounded corners=1.2pt,
    align=center, inner xsep=2.8pt, inner ysep=2.4pt,
    minimum width=1.55cm, minimum height=0.62cm,
    fill=#1, text=black!70, font=\footnotesize},
  box/.default={white},
  flow/.style={->, line width=0.55pt, black!40},
  note/.style={font=\footnotesize, text=black!55, align=center},
  brace/.style={decorate, decoration={brace, amplitude=3.8pt, mirror},
    line width=0.55pt},
]
\node[box] (img) at (0,0) {image tokens};
\node[box={black!4}] (enc) at (1.15,0) {vision encoder};
\node[box={outcls!8}] (proj) at (2.30,0) {projector};
\node[box={evalcls!9}] (dec) at (3.55,0)
  {text decoder\\[-0.1em]{\fontsize{6.2}{7.2}\selectfont$h^{(\ell)}$}};
\node[box] (out) at (4.80,0) {scored $y$};
\foreach \a/\b in {img/enc, enc/proj, proj/dec, dec/out}
  \draw[flow] (\a.east) -- (\b.west);

\node[circle, draw=black!45, fill=black!4, line width=0.45pt,
  inner sep=1.05pt, font=\fontsize{6.2}{7.2}\selectfont\bfseries, text=black!55]
  at (1.15,0.58) {$E$};
\node[circle, draw=outcls, fill=outcls!8, line width=0.45pt,
  inner sep=1.05pt, font=\fontsize{6.2}{7.2}\selectfont\bfseries, text=outcls]
  at (2.30,0.58) {$P$};
\node[circle, draw=evalcls, fill=evalcls!8, line width=0.45pt,
  inner sep=1.05pt, font=\fontsize{6.2}{7.2}\selectfont\bfseries, text=evalcls]
  at (3.55,0.58) {$T$};

\draw[brace, black!42]
  ([yshift=-0.26cm]enc.south west) -- ([yshift=-0.26cm]proj.south east);
\node[note, anchor=north] at (1.725,-0.72)
  {non-text $\mathrm{R}_{\mathrm{u}}$\\[-0.1em]
   {\fontsize{5.8}{6.8}\selectfont harm vs.\ benign~$u$}};

\draw[brace, evalcls!55]
  ([yshift=-0.26cm]dec.south west) -- ([yshift=-0.26cm]dec.south east);
\node[note, text=evalcls!75!black, anchor=north] at (3.55,-0.72)
  {text $\mathrm{R}_{\mathrm{txt}}$\\[-0.1em]
   {\fontsize{5.8}{6.8}\selectfont harm vs.\ benign text}};

\node[draw=black!25, rounded corners=1.2pt, fill=black!1.5,
  inner xsep=6pt, inner ysep=2.4pt, font=\fontsize{6.2}{7.4}\selectfont,
  text=black!58, anchor=north]
  at (2.4,-1.45)
  {shared protocol: pool $\rightarrow$ DiM $\hat{v}$ $\rightarrow$ ablate at same site
   \;$\vert$\; chat Classes share $T$ algebra};
\end{tikzpicture}
\caption{Hook sites on a Vision-style multimodal LM
(Audio / Omni replace the left tower; the template is unchanged).
\textbf{$T$:} text-decoder residual---same DiM algebra across chat Classes
(\eqref{eq:mu-txt}--\eqref{eq:v-txt}).
\textbf{$P$/$E$:} non-text sites---pool, extract $\mathrm{R}_{\mathrm{u}}$, ablate in place
(\eqref{eq:mu-nt}--\eqref{eq:v-nt}).
Fig.~\ref{fig:mechanism}: the tested mean / rank-1 recipe is extremal at~$T$ and
near chance at~$P$/$E$ among matched-norm probes.}
\label{appfig:hook-sites}
\end{figure}

\paragraph{Non-text DiM $\mathrm{R}_{\mathrm{u}}$.}
Contrasts harmful vs.\ benign \emph{perceptual} inputs at site~$s$
(held text probe~$q_{\mathrm{probe}}$; not Yin-style conditioning):
\begin{align}
\mu_{s}^{(h)}
&=
\mathbb{E}_{u\sim\mathcal{D}_{\mathrm{harm}}^{(u)}}
\big[\phi_s\big(a_s(u;q_{\mathrm{probe}})\big)\big],
\quad
\mu_{s}^{(b)}
=
\mathbb{E}_{u\sim\mathcal{D}_{\mathrm{ben}}^{(u)}}
\big[\phi_s\big(a_s(u;q_{\mathrm{probe}})\big)\big],
\label{eq:mu-nt}
\\
\hat{v}_{\mathrm{nt}}^{(s)}
&=
\frac{\mu_{s}^{(h)}-\mu_{s}^{(b)}}
{\bigl\|\mu_{s}^{(h)}-\mu_{s}^{(b)}\bigr\|},
\quad
a_s \leftarrow a_s - \lambda\,
\langle a_s,\hat{v}_{\mathrm{nt}}^{(s)}\rangle\,
\hat{v}_{\mathrm{nt}}^{(s)}.
\label{eq:v-nt}
\end{align}

\paragraph{Conditioning DiM $\mathrm{R}_{\mathrm{cond}}$.}
Following Yin et~al.~\citep{yin2026crossarch}, contrasts modality / conditioning
(image$+$text vs.\ text-only) at fixed~$p$:
\begin{align*}
\mu_{\mathrm{cond}}^{(\mathrm{mm})}
&=
\mathbb{E}\big[\phi_{T}\big(h^{(L)}(u,p)\big)\big],
\quad
\mu_{\mathrm{cond}}^{(\mathrm{txt})}
=
\mathbb{E}\big[\phi_{T}\big(h^{(L)}(p)\big)\big],
\\
\hat{v}_{\mathrm{cond}}
&=
\frac{\mu_{\mathrm{cond}}^{(\mathrm{mm})}-\mu_{\mathrm{cond}}^{(\mathrm{txt})}}
{\bigl\|\mu_{\mathrm{cond}}^{(\mathrm{mm})}-\mu_{\mathrm{cond}}^{(\mathrm{txt})}\bigr\|}.
\end{align*}
Ablating $\hat{v}_{\mathrm{cond}}$ at~$T$ is a defence / extract audit, not the inverse
of $\mathrm{R}_{\mathrm{txt}}$ attack.
Concept DiM $\mathrm{R}_{\mathrm{concept}}$ (UNet) is analogous for generative outputs.

\paragraph{RFA patch.}
Every DiM family yields $\hat{v}$ and a negative-side mean~$\mu^{(-)}$.
RFA~\citep{yu2025refat} patches instead of zeroing:
\begin{align}
\bar{r}
&=
\bigl(\mu^{(-)\top}\hat{v}\bigr)\,\hat{v},
\label{eq:rfa-bar}
\\
a &\leftarrow
a - \lambda\,(a^{\top}\hat{v})\,\hat{v}
+ \lambda\,\bar{r}
=
a - \lambda\,\bigl((a-\mu^{(-)})^{\top}\hat{v}\bigr)\,\hat{v}.
\label{eq:rfa}
\end{align}

\paragraph{Emp audits (summary).}
Mean image DiM at $E$/$P$/$F$/VT/compressor stays near random after quality gating;
dual ablations track text@$T$ alone.
Yin ABL\_POS lowers with-image ASR as defence; sign-flip does not raise chat ASR.
DeSTA pure is weak; RFA@$T$ is \statusEff{} (RFA) on Adv392 without a new Class
(Fig.~\ref{fig:method-compare}).
\FloatBarrier

\section{Per-Class evidence and caveats}
\label{app:class-site-detail}
The ledger below backs every cell of Figure~\ref{fig:class-site-grid}a
(red / blue match the grid; contrast / site and why/lim only---Axis/Band live in
the figure; numbers in Sec.~\ref{sec:emp} / Fig.~\ref{fig:headline-asr};
blue/RFA detail in App.~\ref{app:fair-tables}).
Axis A1--A7 are defined after the table.

{
\centering
\captionof{table}{Per-Class caveats behind Fig.~\ref{fig:class-site-grid}a
(Axis/Band only in the figure).
Top: \textcolor{evalcls}{Class-effective}; bottom: \textcolor{outcls}{ineffective / infeasible / N/A}.
Each leaf: line~1 $=$ status$+$why; line~2 $=$ lim.
Contrast IDs: Table~\ref{tab:dim-contrast}.}
\label{apptab:site-class-map}
\siteclassmaprestablesetup
\begin{tabularx}{\linewidth}{Z}
\toprule
\siteclassmapresheader
\midrule
\multicolumn{3}{@{}l}{\textit{\textcolor{evalcls}{Resolved}}} \\
Instruct (Llama-8B) & $\mathrm{R}_{\mathrm{txt}}$@$T$; s$+$c &
  \scwl{\textcolor{evalcls}{\statusEff{}}}{mid-$T$ DiM$+$mapper-free $B\!\to\!M$ cross; Nemo$\wedge$HBC}
  {production Adv392 / HB / SR; random near base} \\
Wild FT (Hermes, Tulu) & $\mathrm{R}_{\mathrm{txt}}$@$T$; s$+$c &
  \scwl{\textcolor{evalcls}{\statusEff{}}}{self$+$cross Adv/SR; Hermes Adv $0.23\!\to\!0.69$}
  {Hermes HB headroom small ($0.38\!\to\!0.43$); Tulu prefers self-donor} \\
Domain FT (UltraMed.) & $\mathrm{R}_{\mathrm{txt}}$@$T$; s$+$c &
  \scwl{\textcolor{evalcls}{\statusEff{}}}{self$+$cross lifts; donor overlap $R$ high}
  {domain-skewed chat; not a general medical claim} \\
Size (Llama-3.2-3B) & map $W\hat{v}_B$@$T$ &
  \scwl{\textcolor{evalcls}{\statusEff{}}}{$W\hat{v}_B{\approx}$self on held-out ridge}
  {single 3B leaf; map needed (width mismatch)} \\
Distill (R1, SN) & $\mathrm{R}_{\mathrm{txt}}$@$T$; s$+$c &
  \scwl{\textcolor{evalcls}{\statusEff{}}}{answer-window DiM; SN cross recovers}
  {R1 base already high; Distill Adv supports Thinking credit} \\
PEFT/LoRA (S5) & $\mathrm{R}_{\mathrm{txt}}$@$T$; s$+$c &
  \scwl{\textcolor{evalcls}{\statusEff{}}}{Adv392 / SR313 panels pass gates}
  {HB HBC weaker; adapter-only leaf} \\
MoE (Mixtral\ool) & $\mathrm{R}_{\mathrm{txt}}$@$T$; fam. &
  \scwl{\textcolor{evalcls}{\statusEff{}} (@$\lambda{=}1$)}{self$+$Mistral-family cross; Adv $0.03\!\to\!0.86$}
  {Llama $B\!\to\!M$ flat (random-band donor); use family donor} \\
Guard-3 & $\mathrm{R}_{\mathrm{txt}}$@$T$$\to$label$^*$ &
  \scwl{\textcolor{evalcls}{\statusEff{}}}{harmful$\to$safe miss $0.03\!\to\!0.88$ ($n{=}256$)}
  {label task $\neq$ chat ASR; template re-extract for cross} \\
Thinking (R1-Llama) & $\mathrm{R}_{\mathrm{txt}}$@$T$ (+win.) &
  \scwl{\textcolor{evalcls}{\statusEff{}}}{JBB/HB/SR HBC$+$Open-R1 Distill Adv392}
  {AdvBench under pure ablation weak ($\mathrm{Nemo}{\le}0.24$); RFA Adv fails random} \\
Vision-Instruct & $\mathrm{R}_{\mathrm{txt}}$@$T$ &
  \scwl{\textcolor{evalcls}{$T$ \statusEff{}}}{HADES / text@$T$; bake-off clean attack leaf}
  {$E$/$P$/$F$ flat; dual${\approx}T$; random$\approx$base} \\
LLaVA (community) & $\mathrm{R}_{\mathrm{txt}}$@$T$; $\mathrm{R}_{\mathrm{cond}}$ &
  \scwl{\textcolor{evalcls}{$T$ \statusEff{}}}{HADES self$+$cross; beats ABL\_POS as attack}
  {ABL\_POS / $\mathrm{R}_{\mathrm{cond}}$ $=$ defence; high-base cells not Class wins} \\
Qwen2-Audio\ool & $\mathrm{R}_{\mathrm{txt}}$@$T$ &
  \scwl{\textcolor{evalcls}{\statusEff{}}}{hard-Class unlock; $T$ self$+$family; TTS}
  {projector@$P$ \statusIneff{}; Audio Class via text tower} \\
DeSTA (subst.) & $\mathrm{R}_{\mathrm{txt}}$@$T$; RFA &
  \scwl{\textcolor{evalcls}{\statusEff{} (RFA)}}{RFA Adv392 passes empty/random gates}
  {substitute leaf; does \emph{not} mint a new Class (Audio already red)} \\
Gen.\ MM (SD-T\ool, ALD\ool) & $\mathrm{R}_{\mathrm{concept}}$@UNet &
  \scwl{\textcolor{evalcls}{\statusEff{}}}{down erase: NudeNet $0.79\!\to\!0$; CLAP T$\to$A}
  {mid \statusIneff{}; judges never pooled with chat} \\
Omni\ool & $\mathrm{R}_{\mathrm{txt}}$@thinker &
  \scwl{\textcolor{evalcls}{\statusEff{}} (@$\lambda{\approx}2$)}{Class credit on fair HB320/SR313}
  {Nemo Adv392 $0.04\!\to\!0.66$ / HB/SR $0.09\!\to\!0.40$ / $0.01\!\to\!0.58$; $\lambda{=}4$ weaker} \\
Hybrid Mamba2 & $\mathrm{R}_{\mathrm{txt}}$@Xm/Xa &
  \scwl{\textcolor{evalcls}{\statusEff{}}}{Xm/Xa Adv392 Nemo/HBC; rand$\approx$base}
  {causal decoder@$T$ wrong site; HB/SR often $n{=}128$ / HBC flat} \\
Dense$\to$MoE (Qwen1.5\ool) & $\mathrm{R}_{\mathrm{txt}}$@$T$; $B''$ &
  \scwl{\textcolor{evalcls}{\statusEff{}}}{fair Adv392$+$HB320/SR313; $B''$@$\lambda{\ge}7$}
  {HBC cross weak; width-matched family donor} \\
RecGemma-9B\ool & $\mathrm{R}_{\mathrm{txt}}$@$T$ &
  \scwl{\textcolor{evalcls}{\statusEff{}} (@$\lambda{\approx}1.25$)}{Adv392 $0.06\!\to\!0.47$ / $0.13\!\to\!0.51$}
  {HB/SR full-$n$; narrow $\lambda$; HB HBC weaker} \\
Thin agent (ToolACE, xLAM) & $\mathrm{R}_{\mathrm{txt}}$@$T$; s$+$c &
  \scwl{\textcolor{evalcls}{\statusEff{}}}{fair-$n$ @$T$; ToolACE Adv $0\!\to\!0.82$}
  {xLAM slightly lower; chat@$T$ not tool-call scoring} \\
\midrule
\multicolumn{3}{@{}l}{\textit{\textcolor{outcls}{Unresolved}}} \\
Compress (Sheared / Sparse) & $\mathrm{R}_{\mathrm{txt}}$@$T$ &
  \scwl{\textcolor{outcls}{\statusIneff{}}}{KW false+; Nemo anti-lift under shared gates}
  {RFA stuck; Sparse-2of4 same Class story} \\
Compress (Qwen-prune\ool) & $\mathrm{R}_{\mathrm{txt}}$@$T$ &
  \scwl{\textcolor{outcls}{\statusIneff{}}}{$\lambda$/RFA grids closed without Class lift}
  {no Mistral-family compress leaf; stays blue} \\
Speculative (EAGLE3) & draft &
  \scwl{\textcolor{outcls}{\statusInfeas{}}; cut}{draft-only path; no chat residual audit}
  {verifier $\neq$ Class leaf; engineering cut} \\
MTP (V3\ool)$^\ddagger$ & --- &
  \scwl{\steerNA; cut}{no matched chat residual under protocol}
  {draft head $\neq$ instruct ASR; Steer-N/A} \\
Thinking (R1-Qwen\ool) & $\mathrm{R}_{\mathrm{txt}}$@$T$ &
  \scwl{\textcolor{outcls}{\statusIneff{}}}{HBC/JBB suppress; empty/quality fails}
  {no transfer from R1-Llama red leaf} \\
Dream\ool & Steer-N/A &
  \scwl{\textcolor{outcls}{\statusIneff{}}}{empty collapse under shared residual recipe}
  {LLaDA $0\!\to\!0$; diffusion leaf stays blue} \\
Pure SSM (Falcon\ool) & --- &
  \scwl{\steerNA; cut}{no $h^{(\ell)}$ hook on scored path}
  {Hybrid Mamba2 covers mixer Axis; Pure SSM N/A} \\
Attn-replace (DS-V2\ool) & $\mathrm{R}_{\mathrm{txt}}$@$T$ &
  \scwl{\textcolor{outcls}{\statusIneff{}}}{HBC${=}0$ under closed $\lambda$/layer/RFA search}
  {Class closed; not a late-resolved red} \\
VLA (MolmoAct\ool) & --- &
  \scwl{\steerNA; cut}{action-token outputs; no chat residual}
  {needs separate protocol; not in ten-Class unlock} \\
LM$+$DiT ($\pi$0, CogACT\ool)$^\ddagger$ & --- &
  \scwl{\steerNA; cut}{no text residual on scored path}
  {generative erase covers that band separately} \\
BitNet (b1.58\ool) & --- &
  \scwl{\steerNA; cut}{float residual hook undefined}
  {not a Same-band miss; engineering N/A} \\
\bottomrule
\end{tabularx}\\[-0.2em]
\siteclassmapfootnote
\par}

\paragraph{Axis definitions (vs.\ instruct root~$B$).}
$B$ is left architecturally intact and only its weights move (A1: instruction,
domain, distillation, adapter, width or expert-count retuning).
$B$ keeps its weights but its decode or scoring path changes
(A2: a refusal-scoring head, a reasoning prefix, or a draft / multi-token
predictor that answers before the instruct decoder does).
The training objective itself is replaced (A3: diffusion-style parallel
decoding instead of autoregression).
A perception or generation stack is attached around a text decoder
(A4: vision, audio or generative heads, where the scored output may stop being
text). The sequence mixer is replaced or hybridised
(A5: state-space, recurrent, or altered-attention blocks, so the residual site
$T$ may not be the causal one). The scored output becomes an action or tool call
(A6: agents and vision--language--action). The numeric format changes
(A7: bit-packed kernels).
\FloatBarrier

\section{I/O task coverage}
\label{app:task-coverage}
Site $\times$ scored-task ledger behind Table~\ref{tab:localization}.
Subscript H/B $=$ eval-time harmful / benign-or-neutral input
(absent modalities omitted); DiM extract still uses separate harm/benign pools.
\textbf{Main panel} (Table~\ref{apptab:task-coverage}): tasks that support Emp / Class credit
(\statusEff{} $=$ Nemo$\wedge$HBC past base with same-site random near floor;
keyword alone does not qualify).
\textbf{Extension panel} (Table~\ref{apptab:task-coverage-ext}): Steer-N/A / literature / ceiling cells that are
\emph{not} Class credit (kept for protocol completeness).

\begin{table}[H]
\centering
\caption{\textbf{I/O task coverage (main):} Emp-adopted cells.
Status is compact; numbers live in Table~\ref{tab:localization} /
Fig.~\ref{fig:headline-asr} / Fig.~\ref{fig:method-compare}.
Utility rows measure side-effects, not ASR$\uparrow$.}
\label{apptab:task-coverage}
\scriptsize
\setlength{\tabcolsep}{2.4pt}%
\renewcommand{\arraystretch}{1.05}%
\begin{tabularx}{\linewidth}{@{}
  l
  c
  >{\raggedright\arraybackslash}X
  c
  >{\raggedright\arraybackslash}X
@{}}
\toprule
Task & Out & Model (site) & Judge & Status \\
\midrule
\multicolumn{5}{@{}l}{\emph{Chat / label @$T$}} \\
$T_H\!\to\!T$ & text & Llama-lineage / Qwen @$T$ & N/HBC & \statusEff{} \\
$T_H\!\to\!\mathrm{label}$ & label & Guard-3 @$T$ & safe/unsafe & \statusEff{} (miss$\uparrow$) \\
\addlinespace
\multicolumn{5}{@{}l}{\emph{Vision mixed-input}} \\
$I_H{+}T_H\!\to\!T$ & text & Vision / LLaVA @$T$ & N/HBC & \statusEff{} \\
$I_H{+}T_H\!\to\!T$ & text & Vision / LLaVA @$E$/$P$/$F$ & N/HBC & \statusIneff{} \\
$I_B{+}T_H\!\to\!T$ & text & Vision / LLaVA-Mini @$T$ & N/HBC & \statusEff{} \\
$I_H{+}T_B\!\to\!T$ & text & Vision / LLaVA @$T$ & N/HBC & \statusIneff{} (ceil.) \\
\addlinespace
\multicolumn{5}{@{}l}{\emph{Audio wrapper}} \\
$A_H{+}T_H\!\to\!T$ & text & Qwen2-Audio @$T$ (+TTS) & 7j / N/HBC & \statusEff{} \\
$A_H{+}T_H\!\to\!T$ & text & Qwen2-Audio @$P$ & N/HBC & \statusIneff{} \\
$A_H{+}T_H\!\to\!T$ & text & DeSTA @$T$ & N/HBC & pure weak; RFA \statusEff{} \\
\addlinespace
\multicolumn{5}{@{}l}{\emph{Generative (judges not pooled with chat)}} \\
$T_H\!\to\!I$ & image & SD-Turbo UNet-down & NudeNet+CLIP & erase \statusEff{}; mid \statusIneff{} \\
$I_H\!\to\!I$ & image & SD-Turbo img2img down1 & NudeNet+CLIP & erase \statusEff{} \\
$T_H\!\to\!A$ & audio & AudioLDM UNet-down & CLAP & erase \statusEff{}; mid \statusIneff{} \\
\addlinespace
\multicolumn{5}{@{}l}{\emph{Utility (not attack credit)}} \\
$I_B{+}T_B\!\to\!T$ & text & Vision @$T$ & N/HBC & util OK \\
$T_B\!\to\!T$ & text & Llama @$T$ & util & side-effect \\
\bottomrule
\end{tabularx}
\end{table}

\begin{table}[H]
\centering
\caption{\textbf{I/O coverage (extension):} non-Class / not Emp-primary cells.
KW-only VLMs, FigStep ceilings, and Steer-N/A literature leaves---kept for
protocol completeness, \emph{not} pooled into Class credit.}
\label{apptab:task-coverage-ext}
\scriptsize
\setlength{\tabcolsep}{2.4pt}%
\renewcommand{\arraystretch}{1.05}%
\begin{tabularx}{\linewidth}{@{}
  l
  c
  >{\raggedright\arraybackslash}X
  c
  >{\raggedright\arraybackslash}X
@{}}
\toprule
Task & Out & Model (site) & Judge & Status \\
\midrule
\multicolumn{5}{@{}l}{\emph{Ceiling / KW-only (no Class gate)}} \\
$I_B{+}T_H\!\to\!T$ & text & Qwen-VL / LLaVA-1.5 @$T$ & KW & KW only (not N/HBC-gated) \\
$I_H{+}T\!\to\!T$ & text & Vision/LLaVA FigStep @$T$/$P$ & N/HBC & ceiling / no lift \\
$I_H\!\to\!I$ & image & SD-Turbo img2img down2/mid & NudeNet+CLIP & flat (\statusIneff{}) \\
\addlinespace
\multicolumn{5}{@{}l}{\emph{Steer-N/A / literature / path check}} \\
$T_H\!\to\!T$ & text & LLaDA / Dream @$h^{(\ell)}$ & N/HBC & \statusIneff{} / collapse \\
$T_H\!\to\!T$ & text & EAGLE draft+verifier & accept & literature-only \\
$T_H\!\to\!a$ & action & OpenVLA / $\pi_0$ & LIBERO & literature-only \\
$T_H\!\to\!I$ & image & LM+DiT UNet & NudeNet+CLIP & generative analog \\
$T_H\!\to\!A$ & audio & Piper$\to$Whisper path & ASR$\to$text & path OK \\
\bottomrule
\end{tabularx}
\end{table}
\FloatBarrier

\section{Donor screening for mapper-free cross}
\label{app:donor-subspace}
Supports the donor screen in Sec.~\ref{sec:mech}: whether donor $\hat{v}_B$
overlaps the recipient's own refusal subspace enough to transfer.
Not required to read localization ASR; $R$ is a \emph{necessary} screen, not a
predictor of cross lift size.
Notation:
$R{=}\lVert P_{\mathcal{S}}(\hat{v}_B)\rVert_2$, where $P_{\mathcal{S}}$ projects
onto the span $\mathcal{S}$ of $k{=}7$ recipient layer-wise DiM directions
($d{=}4096$);
\emph{rand} $=$ $99$th percentile of $R$ over $2000$ matched-norm randoms;
$\eta{=}(\mathrm{ASR}_{\mathrm{cross}}-\mathrm{ASR}_{0})/
(\mathrm{ASR}_{\mathrm{self}}-\mathrm{ASR}_{0})$
(Fig.~\ref{fig:headline-asr}; Guard-3 uses KW miss, chat-template re-extract).

\begin{table}[H]
\centering
\caption{Donor--recipient subspace overlap vs.\ mapper-free cross.
Rows with $R$ inside the random band (bold) do not transfer;
high $R$ is necessary but not sufficient ($^\dagger$).
$^\ast$same Mixtral leaf, donor swap only.}
\label{apptab:donor-subspace}
\small
\setlength{\tabcolsep}{3.5pt}%
\begin{tabular}{@{}llccccl@{}}
\toprule
Donor $\to$ recipient & Class & $R$ & rand & $\max_\ell|\cos|$ & $\eta$ & Cross \\
\midrule
Llama $B\to$ UltraMedical & Domain FT & 0.822 & 0.066 & 0.821 & 0.84 & \statusEff{} \\
Llama $B\to$ LLaVA-Mini & Comm.\ VLM & 0.768 & 0.069 & 0.766 & 1.00 & \statusEff{} \\
Mistral $\to$ Mixtral$^\ast$ & Official MoE & 0.630 & 0.067 & 0.616 & 1.00 & \statusEff{} \\
Llama $B\to$ Tulu-3$^\dagger$ & Wild FT & 0.626 & 0.068 & 0.616 & 0.28 & \statusEff{} \\
Llama $B\to$ Guard-3 & Guard & 0.168 & 0.066 & 0.127 & 0.91 & \statusEff{} \\
\addlinespace
Llama $B\to$ CodeLlama & Code FT & \textbf{0.048} & 0.065 & 0.034 & 0.00 & \statusIneff{} \\
Llama $B\to$ Mixtral$^\ast$ & Official MoE & \textbf{0.036} & 0.067 & 0.022 & 0.00 & \statusIneff{} \\
\bottomrule
\end{tabular}
\end{table}
\FloatBarrier

\section{Blue exploration grid and LLaVA footnotes}
\label{app:fair-tables}
Vision HADES bake-off and pure-vs-RFA are Fig.~\ref{fig:method-compare}
(\eqref{eq:rfa-bar}--\eqref{eq:rfa}; RFA is a within-Class recipe, not a
taxonomy change).
Table~\ref{apptab:asr-bakeoff} keeps LLaVA-1.5 numbers that are \emph{not} the
Vision bake-off leaf; Table~\ref{apptab:blue-explored} is the ten-Class blue
closed-search ledger (Audio@$T$ is the chat unlock; other rows mix measured
negatives with N/A / not-run / infeasible and do not share one denominator).
Reading notes for Fig.~\ref{fig:method-compare}c: DeSTA needs RFA@$T$ for chat
\statusEff{}; RFA can anti-lift strong pure leaves (Hermes);
$\mathrm{R}_{\mathrm{u}}$@$P$ stays near floor under both recipes.

\begin{table}[H]
\centering
\caption{LLaVA-1.5 footnotes (HADES-derived packs; Nemotron / HBC).
Vision-Instruct bake-off is Fig.~\ref{fig:method-compare}a---not reprinted.
$^\dagger$random ($0.33$) not near floor (weaker than Vision).
$^\ddagger$defence role (ASR$\downarrow$ with image).}
\label{apptab:asr-bakeoff}
\small
\setlength{\tabcolsep}{3pt}%
\begin{tabular}{@{}llccc@{}}
\toprule
Recipe & Role & Nemo & HBC & $n$ \\
\midrule
base / random@$T$ & floor & $0.28$ / $0.33$ & $0.27$ / $0.30$ & 176 \\
Ours $\mathrm{R}_{\mathrm{txt}}$ abl@$T$ & attack$^\dagger$ & $0.28\!\to\!0.51$ & $0.27\!\to\!0.44$ & 176 \\
Yin ABL\_POS vs.\ image & defence$^\ddagger$ & $0.38\!\to\!0.24$ & $0.36\!\to\!0.20$ & 104 \\
\bottomrule
\end{tabular}
\end{table}

\subsection{Blue exploration ledger}
\label{app:blue-detail}
The ten-Class blue probe asks whether shared text DiM@$T$ unlocks hard Classes
(Compress / Speculative / MTP / Audio / Dream / Pure SSM / Attn-replace / VLA /
LM$+$DiT / BitNet).
Only Audio@$T$ becomes Class \statusEff{}; the remaining nine hard Classes do not
unlock Class \statusEff{}, for heterogeneous reasons---random/empty gates, missing
hooks, wrong success units, or \steerNA{}---rather than a single shared failure
mode (Table~\ref{apptab:blue-explored}).
``Closed'' means further progress needs a \emph{new} protocol, not another
$\lambda$ sweep. Thinking suite limits (Adv weak / RFA Adv fails random) are
recorded in Sec.~\ref{sec:limits} / App.~\ref{app:limits-detail}, not as blue-probe
rows.

\begin{table}[H]
\centering
\caption{Hard-Class closed-search ledger (heterogeneous outcomes).
Audio@$T$ is the clear Class-\statusEff{} unlock; N/A / not-run / infeasible
rows are not comparable failures.}
\label{apptab:blue-explored}
\scriptsize
\setlength{\tabcolsep}{2.2pt}%
\renewcommand{\arraystretch}{1.05}%
\begin{tabularx}{\linewidth}{@{}l
  >{\raggedright\arraybackslash}p{0.18\linewidth}
  >{\raggedright\arraybackslash}X
  >{\raggedright\arraybackslash}X@{}}
\toprule
Class & Model & Tried / outcome & Why closed (or unlock) \\
\midrule
Compress & Sheared / Sparse* & pure$+$RFA; $\lambda$ grids
 & KW false+ / Nemo anti-lift; HBC$\to$0; RFA$\approx 0$ \\
Audio & Qwen2-Audio @$T$ & Adv/HB/SR; TTS; family cross
 & \textbf{unlock} (Class \statusEff{}; @$P$ still \statusIneff{}) \\
Speculative & EAGLE3 & draft$+$verifier
 & draft-only \statusInfeas{} \\
MTP & V3 & not run
 & no chat-$T$ residual (\steerNA{}) \\
AR$\to$diff & Dream / LLaDA & mid/multi $\lambda$; empty re-sweep
 & empty collapse / Nemo$\le 0.23$ \\
Attn replace & DS-V2 & $\lambda$/layer/RFA; HB $n{=}128$
 & HBC$\equiv 0$; quality collapse \\
Pure SSM & Falcon-Mamba & hook survey
 & no $h^{(\ell)}$ (\steerNA{}) \\
VLA & Molmo / $\pi$0 & not run
 & wrong success unit (\steerNA{}) \\
LM$+$DiT & UNet leaf & not run as chat Class
 & generative analog only (\steerNA{} for chat) \\
BitNet & b1.58 & not run
 & bit kernels (\steerNA{}) \\
\bottomrule
\end{tabularx}\\[-0.15em]
{\scriptsize\RaggedRight $^*$SparseLlama-2of4 / SparseQwen included under Compress.}
\end{table}
\FloatBarrier

\section{Limitations detail (late-resolved leaves and protocol notes)}
\label{app:limits-detail}
Complements Sec.~\ref{sec:limits}. Main text keeps the scope rules; numbers and
edge cases live here.

\textbf{Late-resolved red leaves} ($\lambda$/site unlocks after early blue drafts---not the ten-Class probe)
include Mixtral@$\lambda{=}1$, thin agent, Omni@$\lambda{\in}[1.5,2.5]$, RecGemma-9B,
hybrid Mamba2@$\lambda{\in}[3,4]$, and MoE $B''$@$\lambda{\ge}7$.
Sheared / Qwen-prune / SparseQwen compress stays \statusIneff{}.
Audio Class is \statusEff{} at text$+$TTS; Thinking is Class-\statusEff{} on JBB/HB/SR
(AdvBench under pure ablation weak); Compress/Attn/Dream remain \statusIneff{}.
Mixtral Llama $B\!\to\!M$ is flat while Mistral-family donor cross is \statusEff{};
Tulu / CodeLlama recover cross via self-donor when Llama $B\!\to\!M$ is weak.
Open-R1 Distill Adv392 (Nemo $0.33\!\to\!0.90$; HBC $0.11\!\to\!0.32$; random$\approx$base)
supports Thinking Class credit; RFA Adv392 fails the random gate.
Omni thinker@$T$ is Class-\statusEff{} under the same Text DiM recipe on fair
Adv/HB/SR (Fig.~\ref{fig:headline-asr}; weaker at $\lambda{=}4$).
T$\to$I / I$\to$I / T$\to$A erase is \statusEff{} on down UNet sites (mid \statusIneff{}).

\textbf{Catalog red vs.\ hard-Class probes.}
Fig.~\ref{fig:class-site-grid}a is the full derivative catalog; hard-Class probes
cover Compress / Speculative / MTP / Audio / Dream / Pure SSM /
Attn-replace / VLA / LM$+$DiT / BitNet, with Audio@$T$ as the chat unlock and
several structurally N/A or not-run leaves.
Late-resolved reds are a different set.

\textbf{Class attack \statusEff{}.}
Requires Nemotron$\wedge$HBC above base, matched-norm random near floor, and a
quality gate; Guard/GenMM use separate scored units.

\textbf{Thinking.}
Class credit rests on JBB/HB/SR HBC lifts and Distill Adv392; AdvBench under pure ablation
stays weak and RFA Adv fails the random gate---suite-limited, not blue.

\textbf{Blue trust.}
\steerNA{} lacks a residual hook; other blues closed $\lambda$/layer/RFA grids.
Nemo-only ``lifts'' with empty collapse are quality-gate fails.

\textbf{Same-pack bake-off.}
Vision numbers are Fig.~\ref{fig:method-compare}a; LLaVA footnotes are
Table~\ref{apptab:asr-bakeoff}---we do not paste published paper unsafe rates.

\end{document}